\documentclass{article} %
\usepackage{iclr2023_conference,times}

\usepackage{amsmath,amsfonts,bm}

\def\eqref#1{equation~\ref{#1}}

\def\1{\bm{1}}

\DeclareMathAlphabet{\mathsfit}{\encodingdefault}{\sfdefault}{m}{sl}
\SetMathAlphabet{\mathsfit}{bold}{\encodingdefault}{\sfdefault}{bx}{n}

\DeclareMathOperator*{\argmax}{arg\,max}
\DeclareMathOperator*{\argmin}{arg\,min}

\usepackage{hyperref}
\usepackage{url}
\usepackage{basecommon}
\usepackage{amsmath}

\usepackage{tabu}
\usepackage{xcolor}
\usepackage{multirow}
\usepackage[font=small,labelfont=bf]{caption}
\usepackage{setspace}
\usepackage{enumitem}
\usepackage{booktabs}
\usepackage{mathtools}
\usepackage{tabularx}
\usepackage{algorithm}
\usepackage{algpseudocode}
\usepackage{graphicx,wrapfig}
\usepackage{inconsolata}

\title{Federated Learning as Variational Inference:\\A Scalable Expectation Propagation Approach}

\author{
\hspace{-4.5mm} $\textbf{Han Guo}^{\dagger \star}$ \hspace{0.65mm} $\textbf{Philip Greengard}^\ddagger$ 
\hspace{0.65mm} $\textbf{Hongyi Wang}^\dagger$  
\hspace{0.65mm} $\textbf{Andrew Gelman}^\ddagger$ 
\hspace{0.65mm} $\textbf{Yoon Kim}^{\star}$ 
\hspace{0.65mm} $\textbf{Eric P. Xing}^{\dagger \diamond}$ \vspace{2mm} \\ 
\normalfont
\textsuperscript{$\dagger$}Carnegie Mellon University, 
\textsuperscript{$\star$}Massachusetts Institute of Technology,  \textsuperscript{$\ddagger$}Columbia University \\
\textsuperscript{$\diamond$}Mohamed bin Zayed University of Artificial Intelligence, Petuum Inc. 
}

\iclrfinalcopy %
\begin{document}
\setlength{\abovedisplayskip}{4pt}
\setlength{\belowdisplayskip}{4pt}
\setlength{\abovedisplayshortskip}{4pt}
\setlength{\belowdisplayshortskip}{4pt}

\maketitle
\vspace{-5mm}
\begin{abstract}
\vspace{-3mm}
The canonical formulation of federated learning treats it as a distributed optimization problem where the model parameters are optimized against a global loss function that decomposes across  client loss functions. A recent alternative formulation instead treats federated learning as a distributed inference problem, where the goal is to infer a global posterior  from partitioned client data \citep{al2020federated}. This paper extends the inference view and describes a variational inference formulation of federated learning where the goal is to find a global variational posterior that well-approximates the true posterior. This naturally motivates an expectation propagation approach to federated learning (FedEP), where approximations to the global posterior are iteratively refined through probabilistic message-passing between the central server and the clients.  We conduct an extensive empirical study across various algorithmic considerations and describe practical strategies for scaling up expectation propagation to the modern federated setting. We apply FedEP on standard federated learning benchmarks and find that it outperforms strong baselines in terms of both convergence speed and accuracy.\footnote{Code: \url{https://github.com/HanGuo97/expectation-propagation}. This work was completed while Han Guo was a visiting student at MIT.}
\vspace{-4mm}
\end{abstract}

\section{Introduction and Background}
\vspace{-2mm}
Many applications of machine learning require training a centralized model over decentralized, heterogeneous, and potentially private datasets.
For example, hospitals may be interested in collaboratively training a model for predictive healthcare, but privacy rules might require each hospital's data to remain local.
{Federated Learning}~\citep[FL,][]{mcmahan2017communication,kairouz2021advances,wang2021field} has emerged as a privacy-preserving training paradigm that does not require clients' private data to leave their local devices. 
FL introduces new challenges on top of classic distributed learning: expensive communication, statistical/hardware heterogeneity, and data privacy~\citep{li2020federated}. 

The canonical formulation of FL treats it as a distributed optimization problem where the model parameters $\btheta$ are trained on $K$ (potentially private) datasets $\mathcal{D} = \bigcup_{k \in [K]} \mathcal{D}_k$, 
\begin{equation*}
\begin{aligned}
\btheta = \argmin\nolimits_{\btheta}\, L(\btheta),
\;\;\;\;\text{where}\;\; L(\btheta) = \sum\nolimits_{k \in [K]} -\log p(\mathcal{D}_k \mid \btheta).
\label{eq:optimization-view}
\end{aligned}
\end{equation*}
Standard distributed optimization algorithms (e.g., data-parallel SGD) are too communication-intensive to be practical under the FL setup.
Federated Averaging~\citep[FedAvg,][]{mcmahan2017communication} reduces communication costs by allowing clients to perform multiple local SGD steps/epochs before the parameter updates are sent back to the central server and aggregated. However, due to client data heterogeneity, more local computations could lead to stale or biased client
updates, and hence sub-optimal behavior~\citep{charles2020outsized,woodworth2020minibatch,wang2020federated}.

An alternative approach  is to consider a \emph{Bayesian} formulation of the FL problem \citep{al2020federated}. Here, we are interested in estimating the posterior of parameters $p(\btheta \mid \mathcal{D})$ given a prior $p(\btheta)$ (such as an improper uniform or a Gaussian prior)  and a collection of client likelihoods $p(\mathcal{D}_k \mid \btheta)$ that are independent given the model parameters,
\begin{equation*}
p(\btheta \mid \mathcal{D}) \propto p(\btheta) \prod\nolimits_{k \in [K]} p(\mathcal{D}_k \mid \btheta).
\label{eq:bayes-rule}
\end{equation*}
In this case the posterior naturally factorizes across partitioned client data, wherein the \emph{global} posterior equates to a multiplicative aggregate of \emph{local} factors (and the prior). However, exact posterior inference is in general intractable for even modestly-sized models and datasets and requires approximate inference techniques. In this paper we turn to variational inference, in effect transforming the  federated optimization problem into a distributed inference problem.
Concretely, we view the solution of federated learning as the mode of a variational (posterior) distribution $q \in \mathcal{Q}$ with some divergence function $D(\cdot \| \cdot)$ (e.g., KL-divergence),
\begin{align}
\btheta = \argmax\nolimits_{\btheta}\, q(\btheta),
\;\;\;\;\text{where}\;\; q(\btheta) =  \argmin\nolimits_{q \in \mathcal{Q}} D\left( p\left(\btheta \mid \mathcal{D}\right) \,\|\, q\left(\btheta\right) \right).
\label{eq:vi-view}
\end{align}
Under this approach, clients use local computation to perform posterior inference (instead of parameter/gradient estimation) in parallel. In exchange, possibly fewer lockstep synchronization and communication steps are required between clients and servers.

One way to operationalize Eq.~\ref{eq:vi-view} is through federated posterior averaging~\citep[FedPA,][]{al2020federated}, where each client independently runs an approximate inference procedure and then sends the local posterior parameters to the server to be multiplicatively aggregated. However, there is no guarantee that independent approximations to local posteriors will lead to a good global approximate posterior.
Motivated by the rich line of  work on variational inference on streaming/partitioned data \citep{broderick2013streaming,vehtari2020expectation}, this work instead considers an \emph{expectation propagation}~\citep[EP,][]{minka2001expectation} approach to FL. In EP, each partition of the data maintains its own local contribution to the global posterior that is iteratively refined through probabilistic message-passing. When applied to FL, this results in an intuitive training scheme where at each round, each client (1) receives the current approximation to the global posterior from the centralized server, (2) carries out local inference to update its local approximation, and (3) sends the refined approximation to the server to be aggregated. Conceptually, this federated learning with expectation propagation (FedEP) approach  extends FedPA by taking into account the current global approximation in step (2).

However, scaling up classic expectation propagation to the modern federated setting is challenging due to the high dimensionality of model parameters and the large number of clients. Indeed, while there is some existing work on expectation propagation-based federated learning \citep{corinzia2019variational,kassab2022fed,ashman2022partitioned}, they typically focus on small models (fewer than 100K parameters) and few clients (at most 100 clients).  In this paper we conduct an extensive empirical study across various algorithmic considerations to scale up expectation propagation to contemporary benchmarks (e.g., models with many millions of parameters and datasets with hundreds of thousands of clients). When applied on top of modern FL benchmarks, our approach outperforms strong FedAvg and FedPA baselines. 
\vspace{-3.5mm}
\section{Federated Learning with Expectation Propagation}
\vspace{-3mm}
The probabilistic view from Eq.~\ref{eq:bayes-rule} motivates an alternative formulation of federated learning based on variational inference.
First observe that the global posterior $p\left(\btheta \mid \mathcal{D}\right)$ given a collection of datasets $\mathcal{D} = \bigcup_{k \in [K]} \mathcal{D}_k$ factorizes as,\vspace{-2.5mm}
\begin{equation*}
\begin{aligned}
p\left(\btheta \mid \mathcal{D}\right)
\propto p(\btheta) \prod_{k = 1}^K p(\mathcal{D}_k \mid \btheta) = \prod_{k=0}^K p_k(\boldsymbol{\theta}),
\end{aligned}
\end{equation*}
where for convenience we define $p_0(\boldsymbol{\theta}) := p(\boldsymbol{\theta})$ to be the prior and further use $p_k(\boldsymbol{\theta}) := p(\mathcal{D}_k \mid \boldsymbol{\theta})$ to refer to the local likelihood associated with $k$-th data partition.
To simplify notation we hereon refer to the global posterior as $p_{\text{global}}(\btheta)$ and drop the conditioning on $\mathcal{D}$. Now consider an approximating global posterior $q_{\text{global}}(\btheta)$ that admits the same factorization as the above, i.e., $q_{\text{global}}(\btheta) \propto \prod_{k=0}^K q_k(\boldsymbol{\theta}).$
Plugging in these terms into Eq.~\ref{eq:vi-view} gives the following objective,\vspace{-0.5mm}
\begin{equation}
\underset{\btheta}{\arg\max}\; \prod_{k=0}^K q_k(\boldsymbol{\theta}),\, \text{ where }
\{q_k(\boldsymbol{\theta})\}_{k=0}^K = \underset{ q_k \in \mathcal{Q}}{\arg\min} \, \infdiv{\prod_{k=0}^K p_k(\boldsymbol{\theta})}{\prod_{k=0}^K q_k(\boldsymbol{\theta})}.
\label{eq:posterior-inference-problem}
\end{equation}
Here $\mathcal{Q}$ is the variational family, which is assumed to be the same for all clients. This  global objective is in general intractable; evaluating $\prod_k p_k(\boldsymbol{\theta})$ requires accessing all clients' data and violates the standard FL assumption. This section presents a probabilistic message-passing algorithm based on expectation propagation \citep[EP,][]{minka2001expectation}.
\vspace{-2mm}
\subsection{Expectation Propagation}
\vspace{-2mm}
\label{subsec:ep-intro}
EP is an iterative algorithm in which an intractable target density $p_{\text{global}}(\btheta)$ is approximated by a tractable density $q_{\text{global}}(\btheta)$ using a collection of localized inference procedures. In EP, each local inference problem is a function of just $p_k$ and the current global estimate, making it appropriate for the FL setting.

Concretely, EP iteratively solves the following problem (either in sequence or parallel),
\begin{equation}
\begin{aligned}
    & q_k^{\text{new}}(\boldsymbol{\theta})
    = \underset{ q \in \mathcal{Q}}{\arg\min} \,
      \infdiv{\underbrace{p_k(\boldsymbol{\theta}) \, {q_{-k}(\btheta)}}_{ \propto\,\,  q_{\backslash k} (\btheta)}}
             {\underbrace{q(\boldsymbol{\theta}) \, {q_{-k}(\btheta)}}_{ \propto\,\,  \widehat{q}_{\backslash k} (\btheta)}},
    & \text{where}\; q_{-k}(\btheta) \propto \frac{q_\text{global}(\boldsymbol{\theta})}{q_k(\boldsymbol{\theta})}.
    \label{eq:ep-kl}
\end{aligned}
\end{equation}
Here $q_{\text{global}}(\boldsymbol{\theta})$ and $q_k^{}(\boldsymbol{\theta})$ are the global/local distributions from the  current iteration. (See Sec.~\ref{subsec:ep-intuitions} for further details).
In the EP literature, $q_{-k}(\btheta)$ is referred to as the \emph{cavity distribution} and $q_{\backslash k} (\btheta)$ and $\widehat{q}_{\backslash k} (\btheta)$ are referred to as the target/approximate \emph{tilted distributions}. EP then uses $q_k^{\text{new}}(\boldsymbol{\theta})$ to derive $q_{\text{global}}^{\text{new}}(\boldsymbol{\theta})$. While the theoretical properties of EP are still not well understood~\citep{minka2001expectation,dehaene2015bounding,dehaene2018expectation}, it has empirically been shown to produce good posterior approximations in many cases~\citep{li2015stochastic,vehtari2020expectation}. When applied to FL, the central server initiates the update by sending the parameters of the current global approximation $q_{\text{global}}(\btheta)$ as messages to the subset of clients $\mathcal{K}$. Upon receiving these messages, each client updates the respective local approximation $q_k^{\text{new}}(\btheta)$ and sends back the changes in parameters as messages,
which is then aggregated by the server.
Algorithms~\ref{alg:ep-general}-\ref{alg:ep-client} illustrate the probabilistic message passing with the Gaussian variational family in more detail.

\vspace{-2mm}
\paragraph{Remark.}
Consider the case where we set $q_{-k}(\btheta) \propto 1$ (i.e., an improper uniform distribution that ignores the current estimate of the global parameters). Then Eq.~\ref{eq:ep-kl} reduces to  federated learning with posterior averaging (FedPA) from \cite{al2020federated}, $q^{\text{new}}_k(\boldsymbol{\theta})
    = {\arg\min_{q \in \mathcal{Q}}} \,
      D\left(p_k(\boldsymbol{\theta})
             \, \Vert \, q(\boldsymbol{\theta})\right)$.
Hence, FedEP improves upon FedPA by taking into account the global parameters and the previous local estimate while deriving the local posterior.\footnote{When the parameters of $q_{\text{global}}(\btheta)$ and $q_k(\btheta)$'s are initialized as improper uniform distributions, the first round (but only the first round) of FedEP and FedPA is identical.}

\begin{figure}[t]
\vspace{-30pt}
\begin{minipage}[t]{.47\textwidth}
\begin{algorithm}[H]
\small
\setstretch{1.30}
\caption{Federated Learning as Inference}
\label{alg:ep-general}
\begin{algorithmic}[1]
\For{round $t = 1, \dots, T$}
    \State Sample a subset of clients $\mathcal{K}$.
    \State \textbf{Broadcast} $q_{\text{global}}(\boldsymbol{\theta})$ to the selected clients.
    \For{each client $k \in \mathcal{K}$ \textbf{in parallel}}
        \State $\Delta q_{k}(\boldsymbol{\theta}) \leftarrow \operatorname{ClientInfer}(q_{\text{global}}(\boldsymbol{\theta}))$
    \EndFor
    \State \textbf{Collect} $\Delta q_{k}(\boldsymbol{\theta})$ from the selected clients.
    \State $q_{\text{global}}(\boldsymbol{\theta}) \leftarrow \operatorname{ServerInfer}(\{ \Delta q_{k}(\boldsymbol{\theta}) \}_{k})$
\EndFor
\State \textbf{Return} $\bmu_{\text{global}}$.
\end{algorithmic}
\end{algorithm}
\vspace{-20pt}
\begin{algorithm}[H]
\small
\setstretch{1.30}
\caption{Approximate Inference: MCMC}
\label{alg:approximate-inference-mcmc}
\begin{algorithmic}[1]
\State \textbf{Input:} $q_{\backslash k}(\btheta; \mathcal{D}_k, \boldeta_{-k}, \bLambda_{-k})$
\State $\mathcal{S}_k \leftarrow \{\}$
\For{$i = 1, \dots, N$}
    \State $\btheta^{(i)}_k \leftarrow \operatorname{SGDEpoch}({-}\log q_{\backslash k},\; \btheta^{(i-1)}_k)$
    \State $\mathcal{S}_k \leftarrow \mathcal{S}_k \cup \btheta^{(i)}_k$
\EndFor
\State $\boldeta_{\backslash k}, \bLambda_{\backslash k} \leftarrow \operatorname{EstimateMoments}(\mathcal{S}_k)$
\State \textbf{Output:} $\widehat{q}_{\backslash k}(\boldsymbol{\theta}; \boldeta_{\backslash k}, \bLambda_{\backslash k})$
\end{algorithmic}
\end{algorithm}
\end{minipage}
\begin{minipage}[t]{.51\textwidth}
\begin{algorithm}[H]
\small
\caption{Gaussian EP: Server Inference}
\label{alg:ep-server}
\begin{algorithmic}[1]
\State \textbf{Receive:} $\{ \Delta q_{k}(\boldsymbol{\theta}; \Delta \boldeta_{k}, \Delta \bLambda_{k}) \}_{k}$
\State {\color{brown} $q_{\text{global}}^{\text{new}} \propto q_{\text{global}} \prod\nolimits_k \left( \Delta q_k \right)^{\delta}$} \Comment{Sec.~\ref{subsec:adaptive-optimization}}
\begin{flalign*}
& \boldeta_{\text {global}} \leftarrow \boldeta_{\text {global}}+\delta \operatorname{ServerOptim}(\sum\nolimits_k \Delta \boldeta_k) &\\
& \bLambda_{\text {global}} \leftarrow \bLambda_{\text {global}}+\delta \operatorname{ServerOptim}(\sum\nolimits_k \Delta \bLambda_k) &
\end{flalign*}
\State \textbf{Send:} $q_{\text{global}}(\boldsymbol{\theta}; \boldeta_{\text {global}}, \bLambda_{\text {global}})$
\end{algorithmic}
\end{algorithm}
\vspace{-20pt}
\begin{algorithm}[H]
\small
\caption{Gaussian EP: Client Inference}
\label{alg:ep-client}
\begin{algorithmic}[1]
\State \textbf{Receive:} $q_{\text{global}}(\boldsymbol{\theta}; \boldeta_{\text{global}}, \bLambda_{\text{global}})$
\State {\color{brown}$q_{-k} \propto q_{\text{global}}/q_k$} \Comment{cavity distribution}
\begin{flalign*}
& \boldeta_{-k} \leftarrow \boldeta_{\text{global}} - \boldeta_{k},
& \bLambda_{-k} \leftarrow \bLambda_{\text{global}} - \bLambda_{k} &
\end{flalign*}
\State {\color{brown} $\widehat{q}_{\backslash k} \approx q_{\backslash k} \propto p_{k} \; q_{-k}$} \Comment{tilted inference (Sec.~\ref{subsubsec:client-inference})}
\begin{flalign*}
& \boldeta_{\backslash k}, \bLambda_{\backslash k} \leftarrow \operatorname{ApproxInference}(q_{\backslash k} \propto p_{k} \; q_{-k} ) &
\end{flalign*}
\State {\color{brown} $\Delta q_{k} \propto \widehat{q}_{\backslash k}/q_{\text{global}}$} \Comment{client deltas (Sec.~\ref{subsec:update-distribution})}
\begin{flalign*}
\small
& \Delta \boldeta_{k} \leftarrow \boldeta_{\backslash k} - \boldeta_{\text{global}},
& \Delta \bLambda_{k} \leftarrow \bLambda_{\backslash k} - \bLambda_{\text{global}} &
\end{flalign*}
\State {\color{brown} $q_k^{\text{new}} \propto q_k \left( \Delta q_k \right)^{\delta}$} \Comment{local update (Sec.~\ref{subsec:adaptive-optimization})}
\begin{flalign*}
&\boldeta_{k} \leftarrow \boldeta_k + \delta \operatorname{ClientOptim}(\Delta\boldeta_{k}) &\\
&\bLambda_{k} \leftarrow \bLambda_k + \delta \operatorname{ClientOptim}(\Delta\bLambda_{k}) &
\end{flalign*}
\State \textbf{Send:} $\Delta q_{k}(\boldsymbol{\theta}; \Delta \boldeta_{k}, \Delta \bLambda_{k})$
\end{algorithmic}
\end{algorithm}
\end{minipage}
\vspace{-10pt}
\end{figure}

\vspace{-2mm}
\subsection{Scalable Expectation Propagation}
\vspace{-2mm}
\label{subsec:scalable-ep}
While federated learning with expectation propagation is conceptually straightforward, scaling up FedEP to modern models and datasets is challenging. For one, the high dimensionality of the parameter space of contemporary models can make local inference difficult even with simple mean-field Gaussian variational families. This is compounded by the fact that classic expectation propagation is  \emph{stateful} and therefore requires that each client always maintains its local contribution to the global posterior. These factors make classic EP potentially an unideal approach in settings where the clients may be resource-constrained and/or the number of clients is large enough that each client is updated only a few times during the course of training. This section discusses various algorithmic consideration when scaling up FedEP to contemporary federated learning benchmarks.

\vspace{-3mm}
\subsubsection{Variational Family}
\vspace{-2mm}
\label{subsec:variational-family}
Following prior work on variational inference in high-dimensional parameter space~\citep{graves2011practical,blundell2015weight,Zhang2019AdvancesIV,osawa2019practical}, we use the mean-field Gaussian variational family for $\mathcal{Q}$, which corresponds to multivariate Gaussian distributions with diagonal covariance. Although non-diagonal extensions are possible (e.g., through shrinkage estimators~\citep{ledoit2004}), we empirically found the diagonal to work well while being simple and communication-efficient. 
For notational simplicity, we use the following two parameterizations of a Gaussian distribution interchangeably,
\begin{equation*}
\begin{aligned}
q(\btheta) = \mathcal{N}\left(\btheta; \bmu, \bSigma\right) = \mathcal{N}\left(\btheta; \boldeta, \bLambda\right),
&&\text{where}\;\;
\bLambda := \bSigma^{-1}, \boldeta := \bSigma^{-1} \bmu.
\end{aligned}
\end{equation*}
Conveniently, both products and quotients of Gaussian distributions---operations commonly used in EP---result in another Gaussian distribution, which simplifies the calculation of the cavity distribution $q_{-k}(\btheta)$ and the global distribution $q_{\text{global}}(\btheta)$.\footnote{Specifically, we have the following identities, \vspace{-2mm}
\begin{equation*}
\begin{aligned}
\mathcal{N}\left(\btheta; \boldeta_1, \bLambda_1\right)  \mathcal{N}\left(\btheta; \boldeta_2, \bLambda_2\right)  \propto \mathcal{N}\left(\btheta; \boldeta_1 + \boldeta_2, \bLambda_1 + \bLambda_2\right), &&
\frac{\mathcal{N}\left(\btheta; \boldeta_1, \bLambda_1\right)}{\mathcal{N}\left(\btheta; \boldeta_2, \bLambda_2\right)} \propto \mathcal{N}\left(\btheta; \boldeta_1 - \boldeta_2, \bLambda_1 - \bLambda_2\right).
\end{aligned}
\end{equation*}
}

\vspace{-2mm}
\subsubsection{Client Inference}
\vspace{-2mm}
\label{subsubsec:client-inference}
At each round of training, each client must estimate $\widehat{q}_{\backslash k}(\btheta)$, its own approximation to the tilted distribution $q_{\backslash k} (\btheta)$ in Eq.~\ref{eq:ep-kl}. We  study various approaches for this estimation procedure. 
\vspace{-3mm}
\paragraph{Stochastic Gradient Markov Chain Monte Carlo (SG-MCMC).}
SG-MCMC~\citep{welling2011bayesian,ma2015complete} uses stochastic gradients to approximately sample from local posteriors. We follow~\cite{al2020federated} and use a simple variant of SGD-based SG-MCMC, where we collect a single sample per epoch to obtain a set of samples $\mathcal{S}_k = \{\btheta_k^{(1)}, \dots, \btheta_k^{(N)}\}$.\footnote{Unlike \cite{al2020federated}, we do not apply Polyak averaging~\citep{mandt2017stochastic,maddox2019simple} as we did not find it to improve results in our case.} The SGD objective in this case is the unnormalized tilted distribution,
\begin{equation*}
\begin{aligned}
\underbrace{- \sum\nolimits_{z \in \mathcal{D}_k} \log p(\boldsymbol{z} \mid \btheta)}_{-\log p_k(\btheta)}
  \underbrace{+ \frac{1}{2} \btheta^{\top} \bLambda_{-k} \btheta
  - \boldeta_{-k}^{\top} \boldsymbol{\theta}}_{-\log q_{-k} (\btheta)},
\end{aligned}
\end{equation*}
which is simply the client negative log likelihood ($-\log p_k(\btheta)$) plus a regularizer that penalizes parameters that have low probability under the cavity distribution ($-\log q_{-k} (\btheta)$). This connection makes it clear that the additional client computation compared to FedAvg (which just minimizes the client negative log-likelihood) is negligible. Given a set of samples $\mathcal{S}_k$ from SG-MCMC, we  estimate the parameters of the tilted distribution $q_{\backslash k}(\btheta)$ with moment matching, i.e., 
\begin{equation*}
\begin{aligned}
q_{\backslash k} (\btheta) = \mathcal{N}(\btheta; \bmu_{\backslash k}, \bSigma_{\backslash k}) && \text{where}\;\;  \bmu_{\backslash k}, \bSigma_{\backslash k} \leftarrow \operatorname{MomentEstimator}(\mathcal{S}_{k}).
\end{aligned}
\end{equation*}
While the mean obtained from $\mathcal{S}_k$ via  averaging empirically worked well, the covariance estimation was sometimes unstable. We next discuss three alternative techniques for  estimating the covariance.

\vspace{-2.5mm}
\paragraph{SG-MCMC with Scaled Identity Covariance.}
Our simplest approach approximates the covariance as a scaled identity matrix with a tunable hyper-parameter $\alpha_{\text{cov}}$, i.e., $\bSigma_{\backslash k} \leftarrow \alpha_{\text{cov}} \boldI.$
This cuts down the communication cost in half since we no longer have to send messages for the covariance parameters.  While extremely simple, we found scaled identity covariance to work well in practice. 

\vspace{-2.5mm}
\paragraph{Laplace Approximation.}
Laplace's method approximates the covariance as the inverse Hessian of the negative log-likelihood at the (possibly approximate) MAP estimate. Since the exact inverse Hessian is intractable, we follow common practice and approximate it with the diagonal Fisher, 
\begin{equation}
\begin{aligned}
\bSigma_{\backslash k} \leftarrow \left(\boldsymbol{H}_k + \bSigma^{-1}_{-k}\right)^{-1},
\;\text{where}\; \boldH_k \approx \underbrace{\operatorname{diag}\left(\mathbb{E}_{x \sim \mathcal{D}_k, \; y \sim p(y \mid x, \btheta)}\left[\bigg(\nabla_\theta \log p (y \mid \btheta, x)\bigg)^2\right]\right)}_{\text{diagonal Fisher approximation}}.
\label{eq:inferece-laplace}
\end{aligned}
\end{equation}
This approach  samples the input $x$ from the data $D_k$ and the output $y$ from the current model $p(y \mid x, \btheta)$, as recommended by~\citet{kunstner2019limitations}. The Fisher approximation requires additional epochs of backpropagation on top of usual SGD (usually 5 in our case), which requires additional client compute.

\vspace{-2.5mm}
\paragraph{Natural Gradient Variational Inference.}
Our final approach uses natural-gradient variational inference~\citep[NGVI,][]{zhang2018noisy,khan2018fast,osawa2019practical}, which incorporates the geometry of the distribution to enable faster convergence. Most existing work on NGVI assume a zero-mean isotropic Gaussian prior. We extend NGVI to work with arbitrary Gaussian priors---necessary for regularizing towards the cavity distribution in FedEP. Specifically, NGVI  iteratively computes the following for $t = 1 \ldots T_{\text{NGVI}}$ and learning rate $\beta_{\text{NGVI}}$,
\begin{equation*}
\begin{aligned}
\bSigma_{\backslash k, t} \leftarrow \left( |\mathcal{D}_k| \bolds_t + \bSigma^{-1}_{-k}  \right)^{-1},\text{where}\;
\bolds_t \leftarrow \beta_{\text{NGVI}} \bolds_{t-1} + \left(1 - \beta_{\text{NGVI}}\right) \mathbb{E}_{\btheta \sim q_{\backslash k, t-1}} \left[ \frac{1}{|\mathcal{D}_k|} \operatorname{Fisher}(\btheta) \right].
\end{aligned}
\end{equation*}
Here $\operatorname{Fisher}(\cdot)$ is the diagonal Fisher approximation in Eq.~\ref{eq:inferece-laplace} but evaluated at a sample of parameters from $q_{\backslash k, t}(\btheta)$, the approximate posterior using the current estimate of $\bSigma_{\backslash k, t}$. We give the exact NGVI update (which is algorithmically similar to the Adam optimizer~\citep{kingma2015adam}) in  Algorithm~\ref{alg:approximate-inference-ngvi} in the appendix.

\vspace{-2mm}
\subsubsection{Adaptive Optimization as Damping}
\label{subsec:adaptive-optimization}
\vspace{-2mm}
Given the approximate tilted distribution $\widehat{q}_{\backslash k}(\btheta)$ and the corresponding parameters $\bmu_{\backslash k}, \bSigma_{\backslash k}$, we can in principle follow the update equation in Eq.~\ref{eq:posterior-inference-problem} to estimate $q_\text{global}^{\text{new}}(\btheta)$. However, adaptive optimizers have been shown to be crucial for scaling federated learning to practical settings~\citep{reddi2020adaptive}, and the vanilla EP update does not immediately lend itself to adaptive updates. This section describes an adaptive extension to EP based on damping, in which we to re-interpret a damped EP update as a  gradient update on the natural parameters, which allows for the use of adaptive optimizers.

Damping performs client updates only partially with step size $\delta$ and is commonly used in parallel EP settings \citep{minka2002expectation,vehtari2020expectation}. Letting
$\Delta q_{k}(\btheta) \propto \widehat{q}_{\backslash k}(\btheta) / q_{\text{global}}(\btheta)$ denote the client ``update'' distribution, we can simplify the update and arrive at the following intuitive form~\citep{vehtari2020expectation} (see Sec.~\ref{subsec:update-distribution} for derivation),
\begin{flalign*}
&\textbf{Client:}
&&q^{\text{new}}_{k}(\btheta)
\propto q_{k}(\btheta) \Big(\Delta q_{k}(\btheta) \Big)^{\delta},
&&\textbf{Server:}
&&q^{\text{new}}_{\text{global}}(\btheta)
\propto q_{\text{global}}(\btheta) \prod_k\Big(\Delta q_{k}(\btheta) \Big)^{\delta}.
&
\end{flalign*}
Recalling that products of Gaussian  distributions yields another Gaussian distribution that simply sums the natural parameters, the damped update for $\boldeta$ is given by,
\begin{flalign*}
&\textbf{Client:}
&&\boldeta_k \leftarrow \boldeta_k + \delta \Delta\boldeta_k,
&&\textbf{Server:}
&&\boldeta_{\text{global}} \leftarrow \boldeta_{\text{global}} + \delta \sum_{k \in \mathcal{K}} \Delta\boldeta_k.
&
\end{flalign*}
(The update on the precision $\bLambda$ is analogous.)
 By re-interpreting the update distribution $\Delta q_k(\btheta; \Delta\boldeta_k, \Delta\bLambda_k)$ as a ``gradient'', we can apply off-the-shelf adaptive optimizers ,
\begin{flalign*}
&\textbf{Client:}
&&\boldeta_k \leftarrow \boldeta_k + \delta \; {\color{red} \operatorname{optim}}(\Delta\boldeta_k),
&&\textbf{Server:}
&&\boldeta_{\text{global}} \leftarrow \boldeta_{\text{global}} + \delta \; {\color{red} \operatorname{optim}}(\sum_{k\in\mathcal{K}} \Delta\boldeta_k).
&
\end{flalign*}
All our FedEP experiments (and the FedAvg and FedPA baselines) employ adaptive optimization.

\vspace{-2mm}
\subsubsection{Stochastic Expectation Propagation for Stateless Clients}
\vspace{-2mm}
\label{subsubsec:stochastic-ep}
Clients are typically assumed to be stateful in the classic formulations of expectation propagation. However, there are scenarios in which stateful clients are infeasible (e.g., memory constraints) or even undesirable (e.g., large number of clients who only participate in a few update rounds, leading to stale messages). We thus additionally experiment with a stateless version of FedEP via stochastic expectation propagation~\citep[SEP,][]{li2015stochastic}. SEP employs direct iterative refinement of a global approximation comprising the prior $p(\btheta)$ and $K$ copies of a \emph{single} approximating factor $\overline{q_{k}}(\btheta)$,
\begin{equation*}
    q_{\text{global}}(\btheta) \propto p(\btheta) \bigg(\overline{q_{k}}(\btheta)\bigg)^{K}.
\end{equation*}
That is, clients are assumed to capture the average effect.
{In practice, FedSEP is implemented in Algorithm~\ref{alg:ep-client} via replacing the cavity update (step $2$) with $q_{-k}(\btheta) \propto q_\text{global}(\boldsymbol{\theta}) / \overline{q_k}(\boldsymbol{\theta})$ and removing the local update (step $5$).}

\vspace{-4mm}
\section{Experiments}
\label{sec:experiments}
\vspace{-2mm}
We empirically study FedEP across various benchmarks. We start with a toy setting in Sec.~\ref{subsec:toy-experiments} where we examine cases where federated posterior average~\citep[FedPA,][]{al2020federated}, which does not take into account global and other clients' approximations during client inference, performs sub-optimally. We then turn to realistic federated learning benchmarks in Sec.~\ref{subsec:benchmark-experiments}, where both the size of the model and the number of clients are much larger than had been previously considered in prior EP-based approaches to federated learning~\citep{corinzia2019variational,kassab2022fed}. Here, we resort to the techniques discussed in Sec.~\ref{subsec:scalable-ep}: approximate inference of the tilted distributions, adaptive optimization, and possibly stateless clients. Finally, we conclude in Sec.~\ref{subsec:analysis-experiments} with an analysis of some of the observations from the benchmark experiments.

\vspace{-2mm}
\subsection{Toy Experiments}
\vspace{-2mm}
\label{subsec:toy-experiments}

\begin{wrapfigure}{r}{0.35\textwidth}
\vspace{-18pt}
 \centering
\includegraphics[width=0.9\linewidth]{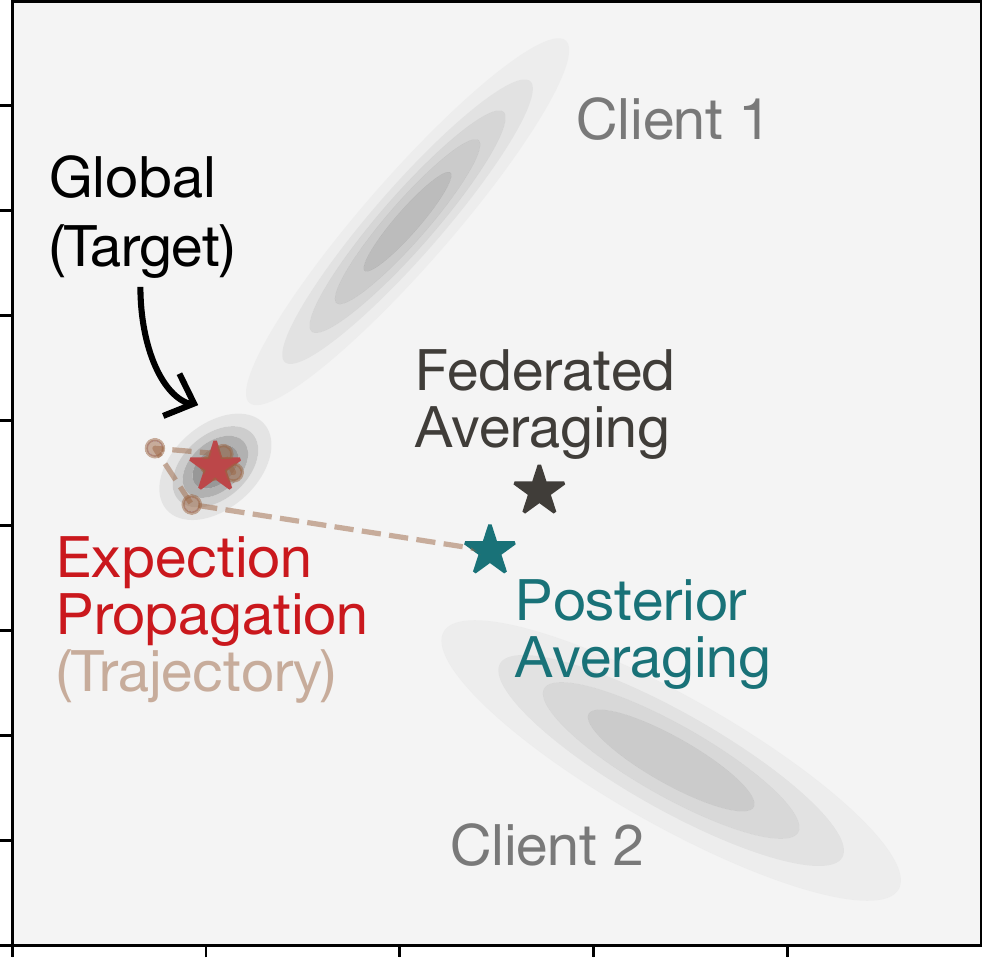}
\vspace{-7pt}
\caption{FedAvg, FedPA, and FedEP on a toy two dimensional dataset with two clients.}
\vspace{-10pt}
\label{fig:toy-experiments-0}
\end{wrapfigure}

We start with a simple toy setting to illustrate the differences between FedPA and FedEP. Here the task is to infer the global mean from two clients, each of which is parameterized as a two-dimensional Gaussian, $p_k(\boldsymbol{\theta}) = \mathcal{N}(\btheta; \bmu_k, \bSigma_k)$ for $k \in \{1, 2\}$. Assuming an improper uniform prior, the global distribution is then also a Gaussian with its posterior mode coinciding with the global mean. We perform exact inference via analytically solving 
$D_{\operatorname{KL}}(q_{\backslash k} \| \widehat{q}_{\backslash k})$, but restrict the variational family to Gaussians with diagonal covariance (i.e., mean-field family). In this case both the FedAvg and FedPA solution can be derived in ``one-shot''.
Fig.~\ref{fig:toy-experiments-0} illustrates a simple case where posterior averaging performs sub-optimally. On the other hand, expectation propagation iteratively refines the approximations toward the globally optimal estimation.

We study this phenomena more systematically by sampling random client distributions,
where the client parameters are sampled from the normal-inverse-Wishart (NIW) distribution,
\begin{equation*}
\begin{aligned}
\boldsymbol{\mu}_k \sim \mathcal{N}\left(\boldsymbol{\mu} \mid \boldsymbol{\mu}_0, \frac{1}{\lambda} \boldsymbol{\Sigma}_k\right),
&&
\boldsymbol{\Sigma}_k \sim \mathcal{W}^{-1}(\boldsymbol{\Sigma} \mid \boldsymbol{\Psi}, \nu).
\end{aligned}
\end{equation*}
Here we set the hyper-prior mean $\bmu_0=\boldsymbol{0}$, degrees of freedom $\nu=7$,  scale $\lambda=0.2$, and sample a random symmetric positive definite matrix for $\boldsymbol{\Psi}$. Table~\ref{table:toy-experiments} shows the average Euclidean distances between the estimated and target global mean for FedAvg, FedPA, and FedEP averaged over $200$ random samples of client distributions. Experimental results demonstrate that iterative message passing in FedEP consistently improves upon the sub-optimal solution from posterior averaging.

\begin{table}[t]
    \vspace{-20pt}
    \centering

    \vspace{-5pt}
    \small
    \begin{tabu} to \linewidth {X[1.7,l] X[1.5,l]X[2.0,l]X[2.6,l]X[4.6,l]}
        \toprule
        \textbf{Dataset}   & \textbf{Model}      & \textbf{Model size}  & \textbf{Clients (train/test)}      & \textbf{Examples per client (train/test)}  \\
        \midrule
        CIFAR-100          & ResNet-18           & $11.2$M               & 500 / 100             & $100_{\pm 0}$ / $100_{\pm 0}$ \\
        StackOverflow      & Linear  & $5.0$M                & 342,477 / 204,088     & $397_{\pm 1279}$ / $81_{\pm 301}$ \\
        EMNIST-62          & CNN       & $1.2$M                & 3,400 / 3,400         & $198_{\pm 77}$ / $23_{\pm 9}$ \\
        \bottomrule
    \end{tabu}
    \vspace{-2mm}
        \caption{Model and dataset statistics. The $\pm$ in ``Examples per client'' client denotes standard deviation.}
        \label{table:datasets}
    \vspace{-5mm}
\end{table}

\begin{table}
\vspace{-20pt}
\begin{minipage}{.495\linewidth}

\scriptsize
\begin{tabular}{wl{1.25cm}|ll|ll}
\toprule
& \multicolumn{2}{c|}{\textbf{Accuracy} ($\%$, $\uparrow$)} & \multicolumn{2}{c}{\textbf{Rounds} ($\#$, $\downarrow$)} \\
\textbf{Method}   & 1000R & 1500R & 45\% & 50\% \\
\midrule
FedPA                & $45.8$ & $48.4$ & $811$ & $-$ \\
FedAvg               & $44.7_{(0.2)}$ & $46.2_{(0.2)}$ & $911_{(86)}$ & $-$ \\
\midrule
FedEP (I)            & $\underline{48.7}_{(0.4)}$ & $\textbf{50.7}_{(0.4)}$ & $\underline{473}_{(17)}$ & $\textbf{1167}_{(107)}$ \\
FedEP (M)            & $\textbf{48.8}_{(0.4)}$ & $\underline{50.4}_{(0.5)}$ & $\textbf{461}_{(13)}$ & $\underline{1240}_{(133)}^\dagger$ \\
FedEP (L)            & $46.5_{(0.4)}$ & $47.7_{(0.3)}$ & $523_{(28)}$ & $-$ \\
FedEP (V)            & $47.8_{(0.5)}$ & $49.6_{(0.6)}$ & $487_{(24)}$ & $1290_{(-)}^\ddagger$ \\
\midrule
FedSEP (I)           & $48.2_{(0.4)}$ & $48.9_{(0.4)}$ & $438_{(9)}$  & $-$ \\
FedSEP (M)           & $48.2_{(0.4)}$ & $48.9_{(0.4)}$ & $438_{(9)}$  & $-$ \\
FedSEP (L)           & $47.2_{(0.4)}$ & $47.8_{(0.4)}$ & $442_{(10)}$ & $-$ \\
FedSEP (V)           & $47.8_{(0.4)}$ & $48.5_{(0.5)}$ & $440_{(10)}$ & $-$ \\
\bottomrule
\end{tabular}
\vspace{-7pt}
\captionof{table}{CIFAR-100 Experiments. Statistics shown are the averages and standard deviations (subscript in brackets) aggregated over $5$ seeds.
We measure the number of rounds to reach certain accuracy thresholds (based on $10$-round running averages) and the best accuracy attained within specific rounds (based on $100$-round running averages).
FedEP and FedSEP refer to the stateful EP and stateless stochastic EP. We use \textbf{I} (Scaled Identity Covariance), \textbf{M} (MCMC), \textbf{L} (Laplace), and \textbf{V} (NGVI) to refer to different inference techniques. $^\dagger$One seed does not reach the threshold. $^\ddagger$Only one seed reaches the threshold.}
\label{table:cifar100-experiments}

\end{minipage}%
\hspace{0.01\linewidth}%
\begin{minipage}{.495\linewidth}
\vspace{-2.5pt}
\scriptsize
\begin{tabular}{wl{1.25cm}|wl{0.9cm}wl{0.9cm}wl{0.9cm}wl{0.9cm}}
\toprule
\textbf {Method} & \textbf{prec.} & \textbf{recall} & \textbf{mi-F1} & \textbf{ma-F1} \\
\midrule
FedPA                & $\underline{74.66}$ & $19.94$ & $30.78$ & $11.63$ \\
FedAvg               & $\textbf{75.20}_{(0.18)}$ & $13.88_{(0.27)}$ & $23.32_{(0.41)}$ & $\;\;8.02_{(0.26)}$ \\
\midrule
FedSEP (I)           & $71.32_{(0.20)}$ & $25.10_{(0.22)}$ & $37.04_{(0.25)}$ & $13.61_{(0.15)}$ \\
FedSEP (M)           & $58.31_{(18.04)}$ & $\;\;8.70_{(1.14)}$ & $14.29_{(2.20)}$ & $\;\;2.70_{(0.33)}$ \\
FedSEP (L)           & $70.98_{(0.18)}$ & $\underline{25.88}_{(0.30)}$ & $\underline{37.80}_{(0.29)}$ & $\underline{13.97}_{(0.19)}$ \\
FedSEP (V)           & $69.51_{(0.35)}$ & $\textbf{28.02}_{(0.20)}$ & $\textbf{39.78}_{(0.25)}$ & $\textbf{15.32}_{(0.08)}$ \\
\bottomrule
\end{tabular}
\vspace{-5pt}
\captionof{table}{StackOverflow Experiments.
Statistics shown are the averages and standard deviations (subscript in brackets) aggregated over $5$ seeds.
We measure the best precision (\textbf{prec.}), \textbf{recall}, micro- and macro-F1 (\textbf{mi/ma-F1}) attained by round $1500$ (based on $100$-round running averages).
}
\label{table:stackoverflow-experiments}

\vspace{7pt}
\small
\begin{tabular*}{\textwidth}{wl{1.25cm}|l@{\extracolsep{\fill}}}
\toprule
\textbf{Method} & \textbf{Euclidean Distance} \\
\midrule
FedAvg & $5.4 \times 10^{-1}$ $\pm\, 4.7 \times 10^{-1}$ \\
FedPA  & $2.6 \times 10^{-1}$ $\pm\, 2.6 \times 10^{-1}$ \\
FedEP  & $1.1 \times 10^{-7}$ $\pm\, 9.8 \times 10^{-8}$ \\
\bottomrule
\end{tabular*}
\vspace{-5pt}
\captionof{table}{Toy Experiments.
Statistics shown are the averages and standard deviations of Euclidean distances between the estimated and target global mean aggregated over $200$ random samples of client distributions.
}
\label{table:toy-experiments}

\end{minipage} 
\vspace{-7mm}
\end{table}

\vspace{-2mm}
\subsection{Benchmarks Experiments}
\vspace{-2mm}
\label{subsec:benchmark-experiments}
We next conduct experiments on a suite of realistic benchmark tasks introduced by~\citet{reddi2020adaptive}. Table~\ref{table:datasets} summarizes the model and raw dataset statistics, which is the same as in~\citet{al2020federated}. We use the dataset preprocessing provided in TensorFlow Federated~\citep[TFF,][]{tff2018}, and implement the models in Jax~\citep{jax2018github,haiku2020github,fedjax2021}.
We compare against both FedAvg with adaptive optimizers  and FedPA.\footnote{\citet{reddi2020adaptive} refer to federated averaging with adaptive server optimizers as FedAdam etc. We refer to this as FedAvg for simplicity.} As in FedPA, we run a few rounds of FedAvg as burn-in before switching to FedEP. We refer the reader to the appendix for the exact experimental setup.

\begin{figure}[t]
    \vspace{-20pt}
    \centering
    \includegraphics[width=0.95\textwidth]{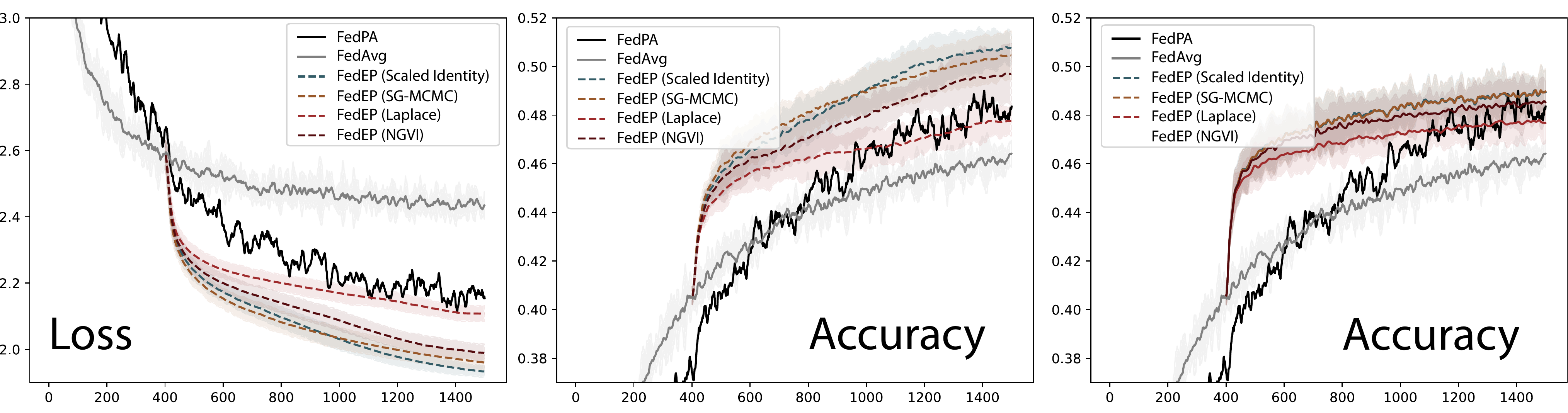}
\vspace{-3mm}
    \caption{CIFAR-100 Experiments. \textbf{Left and Middle:} loss and accuracy of the server as a function of rounds for FedAvg, FedPA, and (stateful) FedEP with various inference techniques. \textbf{Right:} accuracy as a function of rounds for FedAvg, FedPA, and (stateless) FedSEP. The transitions from FedAvg to FedPA, FedEP, and FedSEP happen at round $400$. Lines and shaded regions refer to the averages and $2$ standard deviations over $5$ runs, resp.}
    \label{fig:cifar100-experiments}
\vspace{-4mm}
\end{figure}

For evaluation we consider both convergence speed and final performance. On CIFAR-100 and EMNIST-62, we measure the (1) number of rounds to reach certain accuracy thresholds (based on $10$-round running averages), and (2) the best accuracy attained within specific rounds (based on $100$-round running averages). For StackOverflow, we measure the best precision, recall, micro- and macro-F1 attained by round $1500$ (based on $100$-round running averages).\footnote{TFF by default considers a threshold-based precision and top-5 recall. Our early experiments found that threshold-based metrics correlate better with loss, and use them in StackOverflow experiments.} Due to the size of this dataset, the performance at each round is evaluated on a $10K$ subsample. The evaluation setup is almost exactly the same as in prior work~\citep{reddi2020adaptive,al2020federated}.
Due to space we mainly discuss the CIFAR-100 (``CIFAR'') and StackOverflow Tag Prediction (``StackOverflow'') results in this section and defer the EMNIST-62 (``EMNIST'') results (which are qualitatively similar) to the appendix (Sec.~\ref{subsec:emnist62-experiments}).
\vspace{-2mm}
\paragraph{CIFAR.}
In Table~\ref{fig:cifar100-experiments} and Fig.~\ref{fig:cifar100-experiments} (left, mid), we compare FedAvg, FedPA, and FedEP with various approaches for approximating the clients' tilted distributions (Sec.~\ref{subsubsec:client-inference}). A notable observation is the switch from FedAvg to FedPA/FedEP at the $400$th round, where observe significant increases in performance. Somewhat surprisingly, we find that scaled identity is a simple yet strong baseline. (We conduct further experiments in Sec.~\ref{subsec:analysis-experiments} to analyze this phenomena in greater detail). We next experiment with stochastic EP (FedSEP, Sec.~\ref{subsubsec:stochastic-ep}), a stateless version of FedEP that is more memory-efficient. We find that FedSEP can almost match the performance of full EP despite being much simpler (Fig.~\ref{fig:cifar100-experiments}, right). 
 
\vspace{-2mm}
\paragraph{StackOverflow.}
Experiments on CIFAR study the challenges when scaling FedEP to richly parameterized neural models with millions of parameters. Our StackOverflow experiments are on the other hand intended to investigate whether FedEP can scale to regimes with a large number of clients (hundreds of thousands). Under this setup the number of clients is large enough that the average client will likely only ever participate in a single update round, which renders the stateful version of FedEP meaningless. We thus mainly experiment with the stateless version of FedEP.\footnote{This was also due to the practical difficulty of storing all the clients' distributions.} Table~\ref{table:stackoverflow-experiments} and Fig.~\ref{table:stackoverflow-experiments} (full figure available in the appendix Fig.~\ref{fig:stackoverflow-experiments-full}) show the results comparing the same set of approximate client inference techniques. These experiments demonstrate the scalability of EP to a large number of clients even when we assume clients are stateless.

\begin{figure}[t]
    \centering
    \includegraphics[width=0.97\textwidth]{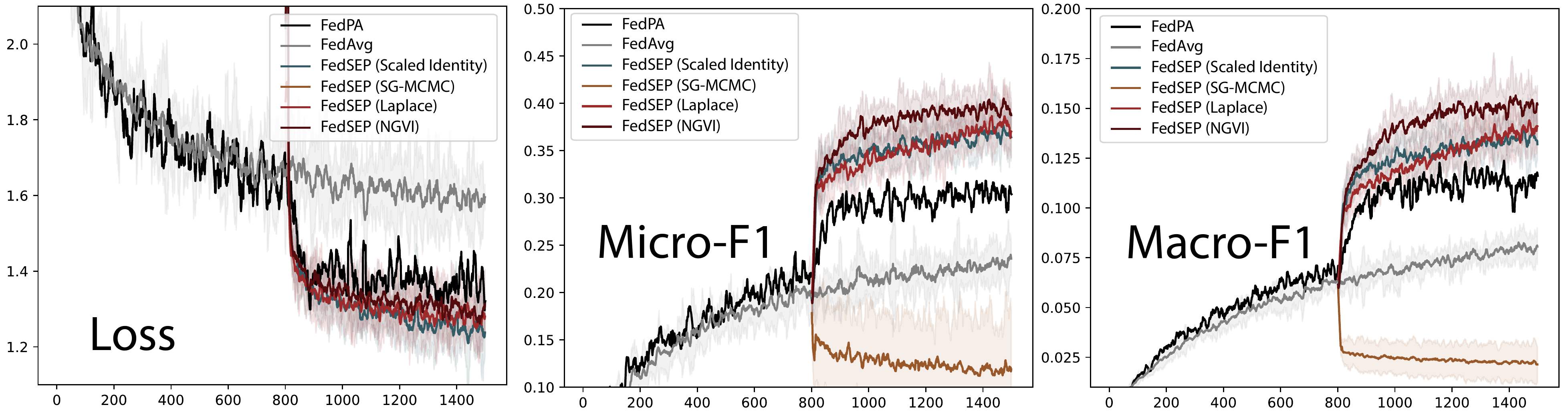}
    \vspace{-3mm}
    \caption{StackOverflow Experiments. Curves represent loss, micro-F1, and macro-F1 of the global parameter estimation as a function of rounds for FedAvg, FedPA, and (stateless) FedSEP with various inference techniques. The transitions from FedAvg to FedPA and FedSEP happen at round $800$. Lines and shaded regions refer to the averages and $2$ standard deviations over $5$ runs, resp.}
    \label{fig:stackoverflow-experiments}
  \vspace{-7mm}
\end{figure}

\begin{figure}[t]
\centering
 \vspace{-7mm}
\includegraphics[width=0.95\textwidth]{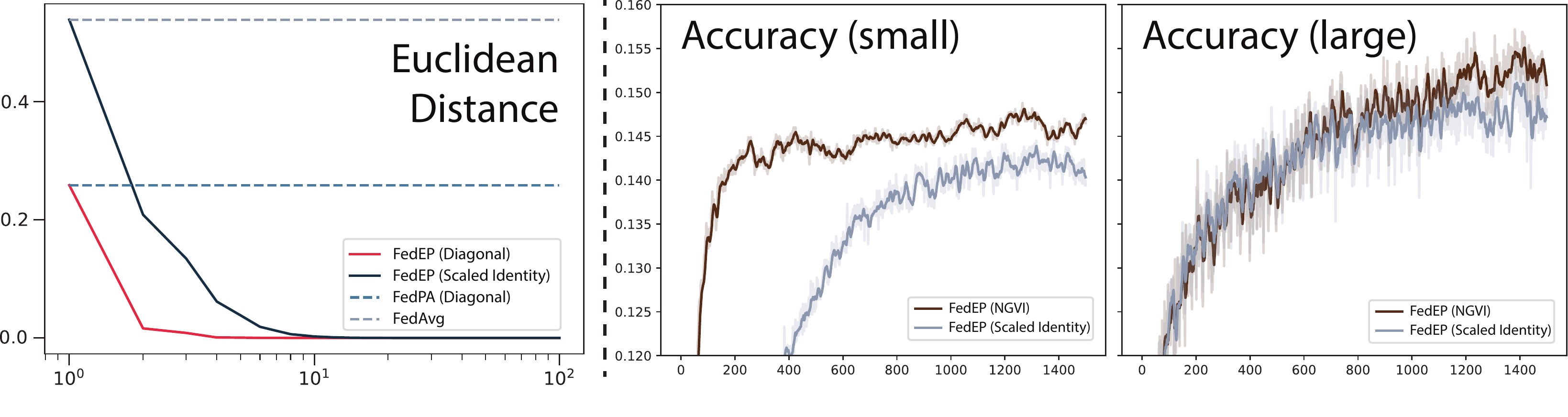}
 \vspace{-5mm}
\caption{Analysis Experiments.
\textbf{Left:} the average Euclidean distances between the estimated and target global mean as a function of rounds in the toy setting.
\textbf{Middle and Right:} accuracy as a function of rounds in the CIFAR-100 setting, with either a (relatively) small model (middle) or large model (right).}
\label{fig:analysis-experiments-combined}
 \vspace{-6mm}
\end{figure}

\vspace{-2mm}
\subsection{Analysis and Discussion}
\vspace{-2mm}
\label{subsec:analysis-experiments}

\paragraph{The Effectiveness of Scaled Identity.}
\label{subsec:analysis-1}
Why does the scaled identity approximation work so well? We investigate this question in the same toy setting as in Sec.~\ref{subsec:toy-experiments}. Fig.~\ref{fig:analysis-experiments-combined} (left) compares the scaled-identity EP with FedEP, FedPA, and FedAvg. Unsurprisingly, this restriction leads to worse performance initially. However, as clients pass messages between each other, scaled-identity EP eventually converges to nearly the same approximation as diagonal EP.

The toy experiments demonstrate the effectiveness of scaled identity in terms of the \emph{final solution}. However, this does not fully explain the  benchmark experiments where we observed scaled identity EP to match more involved variants in terms of \emph{convergence speed}.  We hypothesize that as models grow more complex, the gap between scaled identity and other techniques might decrease due  to the difficulty of obtaining credible estimates of (even diagonal) covariance in high dimensional settings. To test this, we revisit the CIFAR-100 task and compare the following two settings: ``small'' setting which uses a smaller linear model on the PCA'ed features and has $10.1K$ parameters, and a ``large'' setting that uses a linear model on the raw features and has $172.9K$ parameters. For each setting, we conduct experiments with EP using scaled-identity and NGVI and plot the results in Fig.~\ref{fig:analysis-experiments-combined} (right).
We observe that under the ``small'' setting, a more advanced approximate inference technique converges faster than scaled-identity EP, consistent with the toy experiments. As we increase the model size however (``large'' setting), the gap between these two approaches disappears. This indicates that as the model gets more complex, the convergence benefits of more advanced approximate inference decline due to covariance estimation's becoming more difficult.
\begin{wraptable}{r}{.45\textwidth}
\vspace{-3pt}
\scriptsize
\begin{tabular}{wl{1.15cm}|wl{0.7cm}wl{0.7cm}|wl{0.7cm}wl{0.7cm}}
\toprule
& \multicolumn{2}{c|}{\textbf{Accuracy} ($\%$, $\uparrow$)} & \multicolumn{2}{c}{\textbf{ECE-15} ($\%$, $\downarrow$)} \\
\textbf{Method}   & Point Est. & Marg. & Point Est. & Marg. \\
\midrule
FedPA                & $48.1$         & $-$ & $13.6$         & $-$ \\
FedAvg               & $46.6_{(0.7)}$ & $-$ & $19.5_{(0.4)}$ & $-$ \\
\midrule
FedEP (I)            & $50.8_{(0.4)}$ & $49.6_{(0.6)}$ & $4.9 _{(0.3)}$ & $7.9_{(0.2)}$ \\
FedEP (M)            & $50.5_{(0.5)}$ & $50.2_{(0.4)}$ & $5.9 _{(0.5)}$ & $4.6_{(0.4)}$ \\
FedEP (L)            & $47.7_{(0.5)}$ & $47.8_{(0.5)}$ & $8.8 _{(0.4)}$ & $6.6_{(0.4)}$ \\
FedEP (V)            & $49.7_{(0.5)}$ & $49.5_{(0.3)}$ & $5.9 _{(0.4)}$ & $2.2_{(0.5)}$ \\
\midrule
FedSEP (I)           & $49.0_{(0.4)}$ & $48.5_{(0.4)}$ & $10.0_{(0.4)}$ & $3.4_{(0.3)}$ \\
FedSEP (M)           & $48.9_{(0.4)}$ & $48.6_{(0.4)}$ & $10.1_{(0.4)}$ & $3.5_{(0.3)}$ \\
FedSEP (L)           & $47.7_{(0.5)}$ & $47.8_{(0.5)}$ & $9.6 _{(0.6)}$ & $7.2_{(0.6)}$ \\
FedSEP (V)           & $48.5_{(0.4)}$ & $48.7_{(0.4)}$ & $9.3 _{(0.4)}$ & $3.7_{(0.4)}$ \\
\bottomrule
\end{tabular}
\vspace{-7pt}
\captionof{table}{CIFAR-100 Calibration Experiments.
FedEP and FedSEP refer to the stateful EP and stateless stochastic EP. We use \textbf{I} (Scaled Identity Covariance), \textbf{M} (MCMC), \textbf{L} (Laplace), and \textbf{V} (NGVI) to refer to different inference techniques.
}
\vspace{-17pt}
\label{table:cifar100-calibrations}

\end{wraptable}
 \vspace{-6mm}
 \paragraph{Uncertainty Quantification.}
 One motivation for a Bayesian approach is uncertainty quantification. We thus explore whether a Bayesian treatment of federated learning results in models that have better expected calibration error~\citep[ECE,][]{naeini2015obtaining,guo2017calibration}, which is defined as 
$
\operatorname{ECE} = \sum_i^{N_{\text{bins}}} b_i\left|\operatorname{accuracy}_i - \operatorname{confidence}_i\right|.$
Here $\operatorname{accuracy}_i$ is the top-$1$ prediction accuracy in $i$-th bin, $\operatorname{confidence}_i$ is the average confidence of predictions in $i$-th bin, and $b_i$ is the fraction of data points in $i$-th bin. Bins are constructed in a uniform way in the $[0,1]$ range.\footnote{We also experimented with an alternative binning method which puts an equal number of data points in each bin and observed qualitatively similar results.}
We consider accuracy and calibration from the resulting approximate posterior in two ways: (1) point estimation, which uses the final model (i.e., MAP estimate from the approximate posterior) to obtain the output probabilities for each data point, and (2) marginalized estimation, which samples $10$ models from the approximate posterior and averages the output probabilities to obtain the final prediction probability.
In Table~\ref{table:cifar100-calibrations}, we observe that FedEP/FedSEP improves both the accuracy (higher is better) as well as expected calibration error (lower is better), with marginalization sometimes helping.

\vspace{-3mm}
\paragraph{Hyperparameters.} Table~\ref{table:cifar100-hparams} shows the robustness of FedEP w.r.t. various hyperparameters.

\vspace{-3mm}
\paragraph{Limitations.} While FedEP outperforms strong baselines in terms of convergence speed and final accuracy, it has several limitations. The stateful variant requires clients to maintain its current contribution to the global posterior, which increases the clients' memory requirements. The non-scaled-identity approaches also impose additional communication overhead due to the need to communicate the diagonal covariance vector. Further, while SG-MCMC/Scaled-Identity approaches have the same compute cost as FedAvg on the client side, Laplace/NGVI approaches require more compute to estimate the Fisher term. Finally, from a theoretical perspective, while the convergence properties of FedAvg under various assumptions have been extensively studied \citep{li2018fedopt,li2020convergence}, such guarantees for expectation propagation-based approaches remains an open problem.

\vspace{-3.5mm}
\section{Related Work}
\vspace{-3.5mm}
\noindent\textbf{Federated Learning.}
FL is a paradigm for collaborative learning with decentralized private data~\citep{konevcny2016federated,mcmahan2017communication,li2020federated,kairouz2021advances,wang2021field}. Standard approach to FL tackles it as a distributed optimization problem where the global objective is defined by a weighted combination of clients' local objectives~\citep{mohri2019agnostic,li2020federated,reddi2020adaptive,wang2020tackling}. Theoretical analysis has demonstrated that federated optimization exhibits convergence guarantees but only under certain conditions, such as a bounded number of local epochs~\citep{li2020convergence}. Other work has tried to improve the averaging-based aggregations~\cite{yurochkin2019bayesian,wang2020federated}. 
Techniques such as secure aggregation~\citep{bonawitz2017practical,bonawitz2019towards,he2020fedml} and differential privacy~\citep{sun2019can,mcmahan2018learning} have been widely adopted to further improve privacy in FL~\citep{fredrikson2015model}.
Our proposed method is compatible with secure aggregation because it conducts server-side reductions over $\Delta \boldeta_{k}, \Delta \bLambda_{k}$.

\vspace{-1mm}
\noindent\textbf{Expectation Propagation and Approximate Inference.}
This work considers EP as a general technique for passing messages between clients and servers on partitioned data. Here, the cavity distribution ``summarizes'' the effect of inferences from all other partitions and can be used as a prior in the client's local inference. Historically, EP usually refers to a specific choice of divergence function $D_{\text{KL}}(p \| q)$~\citep{minka2001expectation}. This is also known as Variational Message Passing~\citep[VMP,][]{winn2005variational} when $D_{\text{KL}}(q \| p)$ is used instead, and Laplace propagation~\citep[LP,][]{smola2003laplace} when Laplace approximation is used. There have been works that formulate federated learning as a probabilistic inference problem. Most notably,~\citet{al2020federated} formulate FL as a posterior inference problem.~\citet{achituve2021personalized} apply Gaussian processes with deep kernel learning~\citep{wilson2016deep} to personalized FL. Finally, some prior works also consider applying EP to federated learning~\citep{corinzia2019variational,kassab2022fed,ashman2022partitioned}, but mostly on relatively small-scale tasks. In this work, we instead discuss and empirically study various algorithmic considerations to scale up expectation propagation to contemporary benchmarks.

\vspace{-2mm}
\section{Conclusion}
\vspace{-2mm}
This work introduces a probabilistic message-passing algorithm for federated learning based on expectation propagation (FedEP).  Messages (probability distributions) are passed to and from clients to iteratively refine global approximations. To scale up classic expectation propagation to the modern FL setting, we discuss and empirically study various algorithmic considerations, such as choice of variational family, approximate inference techniques, adaptive optimization, and stateful/stateless clients. These enable practical EP algorithms for modern-scale federated learning models and data. 

\paragraph{Reproducibility Statement.}
For experiment details such as the dataset, model, and hyperparameters, we provide detailed descriptions in Sec.~\ref{sec:experiments} as well as Sec.~\ref{subsec:hparams}. We also include in the Appendix additional derivations related to adaptive optimization and damping (Sec.~\ref{subsec:update-distribution}).
\section*{Acknowledgments}
We thank the anonymous reviewers for their comments, and are grateful to Maruan Al-Shedivat for his feedback.
EX is supported by NSF IIS1563887, NSF CCF1629559, NSF IIS1617583, NGA HM04762010002, NIGMS R01GM140467, NSF IIS1955532, NSF CNS2008248, NSF IIS2123952, and NSF BCS2040381. YK acknowledges the support of MIT-IBM Watson AI and Amazon.
\bibliography{citations}

\begin{thebibliography}{54}
\providecommand{\natexlab}[1]{#1}
\providecommand{\url}[1]{\texttt{#1}}
\expandafter\ifx\csname urlstyle\endcsname\relax
  \providecommand{\doi}[1]{doi: #1}\else
  \providecommand{\doi}{doi: \begingroup \urlstyle{rm}\Url}\fi

\bibitem[Achituve et~al.(2021)Achituve, Shamsian, Navon, Chechik, and
  Fetaya]{achituve2021personalized}
Idan Achituve, Aviv Shamsian, Aviv Navon, Gal Chechik, and Ethan Fetaya.
\newblock Personalized federated learning with gaussian processes.
\newblock \emph{Advances in Neural Information Processing Systems},
  34:\penalty0 8392--8406, 2021.

\bibitem[Al-Shedivat et~al.(2021)Al-Shedivat, Gillenwater, Xing, and
  Rostamizadeh]{al2020federated}
Maruan Al-Shedivat, Jennifer Gillenwater, Eric Xing, and Afshin Rostamizadeh.
\newblock Federated learning via posterior averaging: A new perspective and
  practical algorithms.
\newblock In \emph{ICLR}, 2021.

\bibitem[Ashman et~al.(2022)Ashman, Bui, Nguyen, Markou, Weller, Swaroop, and
  Turner]{ashman2022partitioned}
Matthew Ashman, Thang~D Bui, Cuong~V Nguyen, Efstratios Markou, Adrian Weller,
  Siddharth Swaroop, and Richard~E Turner.
\newblock Partitioned variational inference: A framework for probabilistic
  federated learning.
\newblock \emph{arXiv preprint arXiv:2202.12275}, 2022.

\bibitem[Authors(2018)]{tff2018}
The TensorFlow~Federated Authors.
\newblock {TensorFlow Federated}, 12 2018.
\newblock URL \url{https://github.com/tensorflow/federated}.

\bibitem[Blundell et~al.(2015)Blundell, Cornebise, Kavukcuoglu, and
  Wierstra]{blundell2015weight}
Charles Blundell, Julien Cornebise, Koray Kavukcuoglu, and Daan Wierstra.
\newblock Weight uncertainty in neural network.
\newblock In \emph{International conference on machine learning}, pp.\
  1613--1622. PMLR, 2015.

\bibitem[Bonawitz et~al.(2017)Bonawitz, Ivanov, Kreuter, Marcedone, McMahan,
  Patel, Ramage, Segal, and Seth]{bonawitz2017practical}
Keith Bonawitz, Vladimir Ivanov, Ben Kreuter, Antonio Marcedone, H~Brendan
  McMahan, Sarvar Patel, Daniel Ramage, Aaron Segal, and Karn Seth.
\newblock Practical secure aggregation for privacy-preserving machine learning.
\newblock In \emph{proceedings of the 2017 ACM SIGSAC Conference on Computer
  and Communications Security}, pp.\  1175--1191, 2017.

\bibitem[Bonawitz et~al.(2019)Bonawitz, Eichner, Grieskamp, Huba, Ingerman,
  Ivanov, Kiddon, Kone{\v{c}}n{\`y}, Mazzocchi, McMahan,
  et~al.]{bonawitz2019towards}
Keith Bonawitz, Hubert Eichner, Wolfgang Grieskamp, Dzmitry Huba, Alex
  Ingerman, Vladimir Ivanov, Chloe Kiddon, Jakub Kone{\v{c}}n{\`y}, Stefano
  Mazzocchi, Brendan McMahan, et~al.
\newblock Towards federated learning at scale: System design.
\newblock \emph{Proceedings of Machine Learning and Systems}, 1:\penalty0
  374--388, 2019.

\bibitem[Bradbury et~al.(2018)Bradbury, Frostig, Hawkins, Johnson, Leary,
  Maclaurin, Necula, Paszke, Vander{P}las, Wanderman-{M}ilne, and
  Zhang]{jax2018github}
James Bradbury, Roy Frostig, Peter Hawkins, Matthew~James Johnson, Chris Leary,
  Dougal Maclaurin, George Necula, Adam Paszke, Jake Vander{P}las, Skye
  Wanderman-{M}ilne, and Qiao Zhang.
\newblock {JAX}: composable transformations of {P}ython+{N}um{P}y programs,
  2018.
\newblock URL \url{http://github.com/google/jax}.

\bibitem[Broderick et~al.(2013)Broderick, Boyd, Wibisono, Wilson, and
  Jordan]{broderick2013streaming}
Tamara Broderick, Nicholas Boyd, Andre Wibisono, Ashia~C. Wilson, and
  Michael~I. Jordan.
\newblock Streaming variational bayes.
\newblock In Christopher J.~C. Burges, L{\'{e}}on Bottou, Zoubin Ghahramani,
  and Kilian~Q. Weinberger (eds.), \emph{Advances in Neural Information
  Processing Systems}, pp.\  1727--1735, 2013.

\bibitem[Charles \& Kone{\v{c}}n{\`y}(2020)Charles and
  Kone{\v{c}}n{\`y}]{charles2020outsized}
Zachary Charles and Jakub Kone{\v{c}}n{\`y}.
\newblock On the outsized importance of learning rates in local update methods.
\newblock \emph{arXiv preprint arXiv:2007.00878}, 2020.

\bibitem[Corinzia et~al.(2019)Corinzia, Beuret, and
  Buhmann]{corinzia2019variational}
Luca Corinzia, Ami Beuret, and Joachim~M Buhmann.
\newblock Variational federated multi-task learning.
\newblock \emph{arXiv preprint arXiv:1906.06268}, 2019.

\bibitem[Dehaene \& Barthelm{\'e}(2018)Dehaene and
  Barthelm{\'e}]{dehaene2018expectation}
Guillaume Dehaene and Simon Barthelm{\'e}.
\newblock Expectation propagation in the large data limit.
\newblock \emph{Journal of the Royal Statistical Society: Series B (Statistical
  Methodology)}, 80\penalty0 (1):\penalty0 199--217, 2018.

\bibitem[Dehaene \& Barthelm{\'e}(2015)Dehaene and
  Barthelm{\'e}]{dehaene2015bounding}
Guillaume~P Dehaene and Simon Barthelm{\'e}.
\newblock Bounding errors of expectation-propagation.
\newblock \emph{Advances in Neural Information Processing Systems}, 28, 2015.

\bibitem[Fredrikson et~al.(2015)Fredrikson, Jha, and
  Ristenpart]{fredrikson2015model}
Matt Fredrikson, Somesh Jha, and Thomas Ristenpart.
\newblock Model inversion attacks that exploit confidence information and basic
  countermeasures.
\newblock In \emph{Proceedings of the 22nd ACM SIGSAC conference on computer
  and communications security}, pp.\  1322--1333, 2015.

\bibitem[Graves(2011)]{graves2011practical}
Alex Graves.
\newblock Practical variational inference for neural networks.
\newblock \emph{Advances in neural information processing systems}, 24, 2011.

\bibitem[Guo et~al.(2017)Guo, Pleiss, Sun, and Weinberger]{guo2017calibration}
Chuan Guo, Geoff Pleiss, Yu~Sun, and Kilian~Q Weinberger.
\newblock On calibration of modern neural networks.
\newblock In \emph{International Conference on Machine Learning}, pp.\
  1321--1330. PMLR, 2017.

\bibitem[He et~al.(2020)He, Li, So, Zeng, Zhang, Wang, Wang, Vepakomma, Singh,
  Qiu, et~al.]{he2020fedml}
Chaoyang He, Songze Li, Jinhyun So, Xiao Zeng, Mi~Zhang, Hongyi Wang, Xiaoyang
  Wang, Praneeth Vepakomma, Abhishek Singh, Hang Qiu, et~al.
\newblock Fedml: A research library and benchmark for federated machine
  learning.
\newblock \emph{arXiv preprint arXiv:2007.13518}, 2020.

\bibitem[Hennigan et~al.(2020)Hennigan, Cai, Norman, and
  Babuschkin]{haiku2020github}
Tom Hennigan, Trevor Cai, Tamara Norman, and Igor Babuschkin.
\newblock {H}aiku: {S}onnet for {JAX}, 2020.
\newblock URL \url{http://github.com/deepmind/dm-haiku}.

\bibitem[Kairouz et~al.(2021)Kairouz, McMahan, Avent, Bellet, Bennis, Bhagoji,
  Bonawitz, Charles, Cormode, Cummings, et~al.]{kairouz2021advances}
Peter Kairouz, H~Brendan McMahan, Brendan Avent, Aur{\'e}lien Bellet, Mehdi
  Bennis, Arjun~Nitin Bhagoji, Kallista Bonawitz, Zachary Charles, Graham
  Cormode, Rachel Cummings, et~al.
\newblock Advances and open problems in federated learning.
\newblock \emph{Foundations and Trends in Machine Learning}, 14\penalty0
  (1-2):\penalty0 1--210, 2021.

\bibitem[Kassab \& Simeone(2022)Kassab and Simeone]{kassab2022fed}
Rahif Kassab and Osvaldo Simeone.
\newblock Federated generalized bayesian learning via distributed stein
  variational gradient descent.
\newblock \emph{IEEE Transactions on Signal Processing}, 70:\penalty0
  2180--2192, 2022.
\newblock \doi{10.1109/TSP.2022.3168490}.

\bibitem[Khan et~al.(2018)Khan, Nielsen, Tangkaratt, Lin, Gal, and
  Srivastava]{khan2018fast}
Mohammad Khan, Didrik Nielsen, Voot Tangkaratt, Wu~Lin, Yarin Gal, and Akash
  Srivastava.
\newblock Fast and scalable bayesian deep learning by weight-perturbation in
  adam.
\newblock In \emph{International Conference on Machine Learning}, pp.\
  2611--2620. PMLR, 2018.

\bibitem[Kingma \& Ba(2015)Kingma and Ba]{kingma2015adam}
Diederik~P Kingma and Jimmy Ba.
\newblock Adam: A method for stochastic optimization.
\newblock In \emph{ICLR}, 2015.

\bibitem[Kone{\v{c}}n{\`y} et~al.(2016)Kone{\v{c}}n{\`y}, McMahan, Ramage, and
  Richt{\'a}rik]{konevcny2016federated}
Jakub Kone{\v{c}}n{\`y}, H~Brendan McMahan, Daniel Ramage, and Peter
  Richt{\'a}rik.
\newblock Federated optimization: Distributed machine learning for on-device
  intelligence.
\newblock \emph{arXiv preprint arXiv:1610.02527}, 2016.

\bibitem[Kunstner et~al.(2019)Kunstner, Hennig, and
  Balles]{kunstner2019limitations}
Frederik Kunstner, Philipp Hennig, and Lukas Balles.
\newblock Limitations of the empirical {Fisher} approximation for natural
  gradient descent.
\newblock \emph{Advances in Neural Information Processing Systems}, 32, 2019.

\bibitem[Ledoit \& Wolf(2004)Ledoit and Wolf]{ledoit2004}
Olivier Ledoit and Michael Wolf.
\newblock A well-conditioned estimator for large-dimensional covariance
  matrices.
\newblock \emph{Journal of Multivariate Analysis}, 88\penalty0 (2):\penalty0
  365--411, 2004.

\bibitem[Li et~al.(2018)Li, Sahu, Zaheer, Sanjabi, Talwalkar, and
  Smith]{li2018fedopt}
Tian Li, Anit~Kumar Sahu, Manzil Zaheer, Maziar Sanjabi, Ameet Talwalkar, and
  Virginia Smith.
\newblock Federated optimization in heterogeneous networks.
\newblock \emph{arXiv preprint arXiv:1812.06127}, 2018.

\bibitem[Li et~al.(2020{\natexlab{a}})Li, Sahu, Talwalkar, and
  Smith]{li2020federated}
Tian Li, Anit~Kumar Sahu, Ameet Talwalkar, and Virginia Smith.
\newblock Federated learning: Challenges, methods, and future directions.
\newblock \emph{IEEE Signal Processing Magazine}, 37\penalty0 (3):\penalty0
  50--60, 2020{\natexlab{a}}.

\bibitem[Li et~al.(2020{\natexlab{b}})Li, Huang, Yang, Wang, and
  Zhang]{li2020convergence}
Xiang Li, Kaixuan Huang, Wenhao Yang, Shusen Wang, and Zhihua Zhang.
\newblock On the convergence of fedavg on non-iid data.
\newblock In \emph{International Conference on Learning Representations},
  2020{\natexlab{b}}.

\bibitem[Li et~al.(2015)Li, Hern{\'a}ndez-Lobato, and Turner]{li2015stochastic}
Yingzhen Li, Jos{\'e}~Miguel Hern{\'a}ndez-Lobato, and Richard~E Turner.
\newblock Stochastic expectation propagation.
\newblock \emph{NeurIPS}, 2015.

\bibitem[Ma et~al.(2015)Ma, Chen, and Fox]{ma2015complete}
Yi-An Ma, Tianqi Chen, and Emily Fox.
\newblock A complete recipe for stochastic gradient {MCMC}.
\newblock \emph{Advances in Neural Information Processing Systems}, 28, 2015.

\bibitem[Maddox et~al.(2019)Maddox, Izmailov, Garipov, Vetrov, and
  Wilson]{maddox2019simple}
Wesley~J Maddox, Pavel Izmailov, Timur Garipov, Dmitry~P Vetrov, and
  Andrew~Gordon Wilson.
\newblock A simple baseline for bayesian uncertainty in deep learning.
\newblock \emph{Advances in Neural Information Processing Systems}, 32, 2019.

\bibitem[Mandt et~al.(2017)Mandt, Hoffman, and Blei]{mandt2017stochastic}
Stephan Mandt, Matthew~D Hoffman, and David~M Blei.
\newblock Stochastic gradient descent as approximate bayesian inference.
\newblock \emph{arXiv preprint arXiv:1704.04289}, 2017.

\bibitem[McMahan et~al.(2017)McMahan, Moore, Ramage, Hampson, and
  y~Arcas]{mcmahan2017communication}
Brendan McMahan, Eider Moore, Daniel Ramage, Seth Hampson, and Blaise~Aguera
  y~Arcas.
\newblock Communication-efficient learning of deep networks from decentralized
  data.
\newblock In \emph{Artificial intelligence and statistics}, pp.\  1273--1282.
  PMLR, 2017.

\bibitem[McMahan et~al.(2018)McMahan, Ramage, Talwar, and
  Zhang]{mcmahan2018learning}
H~Brendan McMahan, Daniel Ramage, Kunal Talwar, and Li~Zhang.
\newblock Learning differentially private recurrent language models.
\newblock In \emph{International Conference on Learning Representations}, 2018.

\bibitem[Minka \& Lafferty(2002)Minka and Lafferty]{minka2002expectation}
Thomas Minka and John Lafferty.
\newblock Expectation-propagation for the generative aspect model.
\newblock In \emph{Proceedings of the Eighteenth conference on Uncertainty in
  artificial intelligence}, pp.\  352--359, 2002.

\bibitem[Minka(2001)]{minka2001expectation}
Thomas~P Minka.
\newblock Expectation propagation for approximate bayesian inference.
\newblock In \emph{UAI}, 2001.

\bibitem[Mohri et~al.(2019)Mohri, Sivek, and Suresh]{mohri2019agnostic}
Mehryar Mohri, Gary Sivek, and Ananda~Theertha Suresh.
\newblock Agnostic federated learning.
\newblock In \emph{International Conference on Machine Learning}, pp.\
  4615--4625. PMLR, 2019.

\bibitem[Naeini et~al.(2015)Naeini, Cooper, and
  Hauskrecht]{naeini2015obtaining}
Mahdi~Pakdaman Naeini, Gregory Cooper, and Milos Hauskrecht.
\newblock Obtaining well calibrated probabilities using bayesian binning.
\newblock In \emph{Twenty-Ninth AAAI Conference on Artificial Intelligence},
  2015.

\bibitem[Osawa et~al.(2019)Osawa, Swaroop, Khan, Jain, Eschenhagen, Turner, and
  Yokota]{osawa2019practical}
Kazuki Osawa, Siddharth Swaroop, Mohammad Emtiyaz~E Khan, Anirudh Jain, Runa
  Eschenhagen, Richard~E Turner, and Rio Yokota.
\newblock Practical deep learning with {Bayesian} principles.
\newblock \emph{Advances in neural information processing systems}, 32, 2019.

\bibitem[Reddi et~al.(2020)Reddi, Charles, Zaheer, Garrett, Rush,
  Kone{\v{c}}n{\`y}, Kumar, and McMahan]{reddi2020adaptive}
Sashank~J Reddi, Zachary Charles, Manzil Zaheer, Zachary Garrett, Keith Rush,
  Jakub Kone{\v{c}}n{\`y}, Sanjiv Kumar, and Hugh~Brendan McMahan.
\newblock Adaptive federated optimization.
\newblock In \emph{International Conference on Learning Representations}, 2020.

\bibitem[Ro et~al.(2021)Ro, Suresh, and Wu]{fedjax2021}
Jae~Hun Ro, Ananda~Theertha Suresh, and Ke~Wu.
\newblock {F}ed{JAX}: Federated learning simulation with {JAX}.
\newblock \emph{arXiv preprint arXiv:2108.02117}, 2021.

\bibitem[Smola et~al.(2003)Smola, Vishwanathan, and Eskin]{smola2003laplace}
Alex Smola, S.v.n. Vishwanathan, and Eleazar Eskin.
\newblock Laplace propagation.
\newblock In \emph{Advances in Neural Information Processing Systems},
  volume~16, 2003.

\bibitem[Sun et~al.(2019)Sun, Kairouz, Suresh, and McMahan]{sun2019can}
Ziteng Sun, Peter Kairouz, Ananda~Theertha Suresh, and H~Brendan McMahan.
\newblock Can you really backdoor federated learning?
\newblock \emph{arXiv preprint arXiv:1911.07963}, 2019.

\bibitem[Vehtari et~al.(2020)Vehtari, Gelman, Sivula, Jyl{\"a}nki, Tran, Sahai,
  Blomstedt, Cunningham, Schiminovich, and Robert]{vehtari2020expectation}
Aki Vehtari, Andrew Gelman, Tuomas Sivula, Pasi Jyl{\"a}nki, Dustin Tran,
  Swupnil Sahai, Paul Blomstedt, John~P Cunningham, David Schiminovich, and
  Christian~P Robert.
\newblock Expectation propagation as a way of life: A framework for {Bayesian}
  inference on partitioned data.
\newblock \emph{JMLR}, 2020.

\bibitem[Wang et~al.(2020{\natexlab{a}})Wang, Yurochkin, Sun, Papailiopoulos,
  and Khazaeni]{wang2020federated}
Hongyi Wang, Mikhail Yurochkin, Yuekai Sun, Dimitris Papailiopoulos, and
  Yasaman Khazaeni.
\newblock Federated learning with matched averaging.
\newblock In \emph{International Conference on Learning Representations},
  2020{\natexlab{a}}.

\bibitem[Wang et~al.(2020{\natexlab{b}})Wang, Liu, Liang, Joshi, and
  Poor]{wang2020tackling}
Jianyu Wang, Qinghua Liu, Hao Liang, Gauri Joshi, and H~Vincent Poor.
\newblock Tackling the objective inconsistency problem in heterogeneous
  federated optimization.
\newblock \emph{Advances in neural information processing systems},
  33:\penalty0 7611--7623, 2020{\natexlab{b}}.

\bibitem[Wang et~al.(2021)Wang, Charles, Xu, Joshi, McMahan, Al-Shedivat,
  Andrew, Avestimehr, Daly, Data, et~al.]{wang2021field}
Jianyu Wang, Zachary Charles, Zheng Xu, Gauri Joshi, H~Brendan McMahan, Maruan
  Al-Shedivat, Galen Andrew, Salman Avestimehr, Katharine Daly, Deepesh Data,
  et~al.
\newblock A field guide to federated optimization.
\newblock \emph{arXiv preprint arXiv:2107.06917}, 2021.

\bibitem[Welling \& Teh(2011)Welling and Teh]{welling2011bayesian}
Max Welling and Yee~W Teh.
\newblock Bayesian learning via stochastic gradient {Langevin} dynamics.
\newblock In \emph{Proceedings of the 28th International Conference on Machine
  Learning (ICML-11)}, pp.\  681--688. Citeseer, 2011.

\bibitem[Wilson et~al.(2016)Wilson, Hu, Salakhutdinov, and
  Xing]{wilson2016deep}
Andrew~Gordon Wilson, Zhiting Hu, Ruslan Salakhutdinov, and Eric~P Xing.
\newblock Deep kernel learning.
\newblock In \emph{Artificial intelligence and statistics}, pp.\  370--378.
  PMLR, 2016.

\bibitem[Winn et~al.(2005)Winn, Bishop, and Jaakkola]{winn2005variational}
John Winn, Christopher~M Bishop, and Tommi Jaakkola.
\newblock Variational message passing.
\newblock \emph{Journal of Machine Learning Research}, 6\penalty0 (4), 2005.

\bibitem[Woodworth et~al.(2020)Woodworth, Patel, and
  Srebro]{woodworth2020minibatch}
Blake~E Woodworth, Kumar~Kshitij Patel, and Nati Srebro.
\newblock Minibatch vs local sgd for heterogeneous distributed learning.
\newblock \emph{Advances in Neural Information Processing Systems},
  33:\penalty0 6281--6292, 2020.

\bibitem[Yurochkin et~al.(2019)Yurochkin, Agarwal, Ghosh, Greenewald, Hoang,
  and Khazaeni]{yurochkin2019bayesian}
Mikhail Yurochkin, Mayank Agarwal, Soumya Ghosh, Kristjan Greenewald, Nghia
  Hoang, and Yasaman Khazaeni.
\newblock Bayesian nonparametric federated learning of neural networks.
\newblock In \emph{International Conference on Machine Learning}, pp.\
  7252--7261. PMLR, 2019.

\bibitem[Zhang et~al.(2019)Zhang, B{\"u}tepage, Kjellstr{\"o}m, and
  Mandt]{Zhang2019AdvancesIV}
Cheng Zhang, Judith B{\"u}tepage, Hedvig Kjellstr{\"o}m, and Stephan Mandt.
\newblock Advances in variational inference.
\newblock \emph{IEEE Transactions on Pattern Analysis and Machine
  Intelligence}, 41:\penalty0 2008--2026, 2019.

\bibitem[Zhang et~al.(2018)Zhang, Sun, Duvenaud, and Grosse]{zhang2018noisy}
Guodong Zhang, Shengyang Sun, David Duvenaud, and Roger Grosse.
\newblock Noisy natural gradient as variational inference.
\newblock In \emph{International Conference on Machine Learning}, pp.\
  5852--5861. PMLR, 2018.

\end{thebibliography}
\bibliographystyle{iclr2023_conference}

\newpage
\appendix
\section{Appendix}
\subsection{Damped Client and Server Updates}
\label{subsec:update-distribution}
To simplify the notations, observe that Eq.~\ref{eq:ep-kl} could be re-written in the following way,
\begin{equation*}
\begin{aligned}
& q_k^{\text{new}}(\boldsymbol{\theta})
\propto \frac{\widehat{q}_{\backslash k}(\boldsymbol{\theta})}{q_{-k}(\boldsymbol{\theta})},
& \text{where}\; \widehat{q}_{\backslash k}(\boldsymbol{\theta})
= \underset{ \widehat{q}_{\backslash k} \in \mathcal{Q}}{\arg\min} \,
  \infdiv{p_k(\boldsymbol{\theta}) \, {q_{-k}(\btheta)}}
         {\widehat{q}_{\backslash k} (\btheta)}.
\end{aligned}
\end{equation*}
A partially damped client update could be carried out by,
\begin{equation*}
    \begin{aligned}
        \textbf{Client:}
        &&q^{\text{new}}_{k}(\btheta)
        &\propto \bigg(q_{k}(\btheta) \bigg)^{1 - \delta} \bigg(\frac{\widehat{q}_{\backslash k}(\btheta)}{q_{-k}(\btheta)}\bigg)^{\delta}  \\
        &&&\propto \bigg(q_{k}(\btheta) \bigg)^{1 - \delta} \bigg(\frac{\widehat{q}_{\backslash k}(\btheta)}{q_{\text{global}}(\btheta) / q_{k}(\btheta)}\bigg)^{\delta}  \\
        &&&\propto \bigg(q_{k}(\btheta) \bigg)^{1 - \delta} \bigg(q_{k}(\btheta)\bigg)^{\delta} \bigg(\frac{\widehat{q}_{\backslash k}(\btheta)}{q_{\text{global}}(\btheta)}\bigg)^{\delta}  \\
        &&&\propto q_{k}(\btheta) \bigg(\frac{\widehat{q}_{\backslash k}(\btheta)}{q_{\text{global}}(\btheta)}\bigg)^{\delta}  \\
        &&&\propto q_{k}(\btheta) \bigg(\Delta q_{k}(\btheta) \bigg)^{\delta},  \\
        \text{where we define} &&\Delta q_{k}(\btheta) &\propto \frac{\widehat{q}_{\backslash k}(\btheta)}{q_{\text{global}}(\btheta)}.
    \end{aligned}
\end{equation*}
Similarly, (damped) server updates could be written as the following,
\begin{equation*}
    \begin{aligned}
        \textbf{Server:}
        &&q^{\text{new}}_{\text{global}}(\btheta)
        &\propto \prod_k  q^{\text{new}}_{k}(\btheta) \\
        &&&\propto \prod_k  q_{k}(\btheta) \bigg(\Delta q_{k}(\btheta) \bigg)^{\delta} \\
        &&&\propto \bigg[\prod_k  q_{k}(\btheta) \bigg] \bigg[\prod_k\bigg(\Delta q_{k}(\btheta) \bigg)^{\delta}\;\bigg] \\
        &&&\propto q_{\text{global}}(\btheta) \prod_k\bigg(\Delta q_{k}(\btheta) \bigg)^{\delta}. \\
    \end{aligned}
\end{equation*}

\subsection{Expectation Propagation (Extended)}
\label{subsec:ep-intuitions}
Expectation propagation (EP)~\citet{minka2001expectation,vehtari2020expectation} constructs a posterior approximation through iterating local computations that refine factors that approximate the posterior contribution from each client. In this spirit, we would ideally like to solve the following localized version of Eq.~\ref{eq:posterior-inference-problem}, where we replace one of the factors with its corresponding approximating factor,
\begin{equation*}
\begin{aligned}
    & q_k^{\text{new}}(\boldsymbol{\theta})
    = \underset{ q \in \mathcal{Q}}{\arg\min} \,
      \infdiv{p_k(\boldsymbol{\theta}) \, {\color{red} p_{-k}(\btheta)}}
             {q(\boldsymbol{\theta}) \, {\color{red} p_{-k}(\btheta)}},
    & \text{where}\; p_{-k}(\btheta) \propto \frac{p_\text{global}(\boldsymbol{\theta})}{p_k(\boldsymbol{\theta})}.
\end{aligned}
\end{equation*}
Unfortunately, the right-hand side of the divergence is the intractable posterior we would like to approximate in the first place. Instead, EP solves the following problem (Eq.~\ref{eq:ep-kl}),
\begin{equation*}
\begin{aligned}
    & q_k^{\text{new}}(\boldsymbol{\theta})
    = \underset{ q \in \mathcal{Q}}{\arg\min} \,
      \infdiv{p_k(\boldsymbol{\theta}) \, {\color{red} q_{-k}(\btheta)}}
             {q(\boldsymbol{\theta}) \, {\color{red} q_{-k}(\btheta)}},
    & \text{where}\; q_{-k}(\btheta) \propto \frac{q_\text{global}(\boldsymbol{\theta})}{q_k(\boldsymbol{\theta})}.
\end{aligned}
\end{equation*}

\subsection{Additional Experiments and Details}
\begin{figure}[t]
    \vspace{-12pt}
    \centering
    \includegraphics[width=0.99\textwidth]{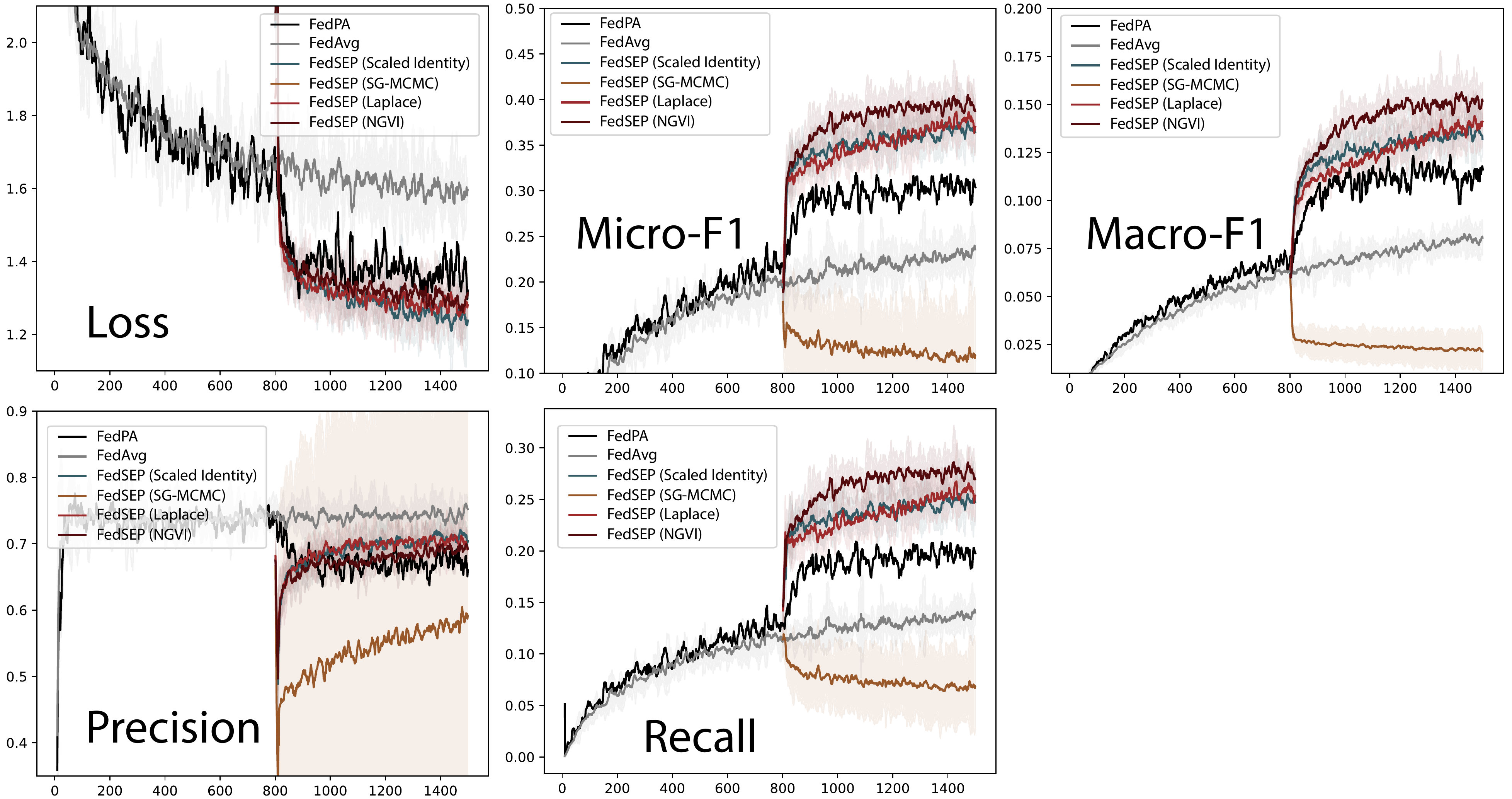}
    \caption{Extended StackOverflow Visualizations.}
    \label{fig:stackoverflow-experiments-full}
\end{figure}
\paragraph{StackOverflow.}
Please see Fig.~\ref{fig:stackoverflow-experiments-full} for additional visualizations.

\begin{figure}[t]
\centering
\includegraphics[width=0.7\textwidth]{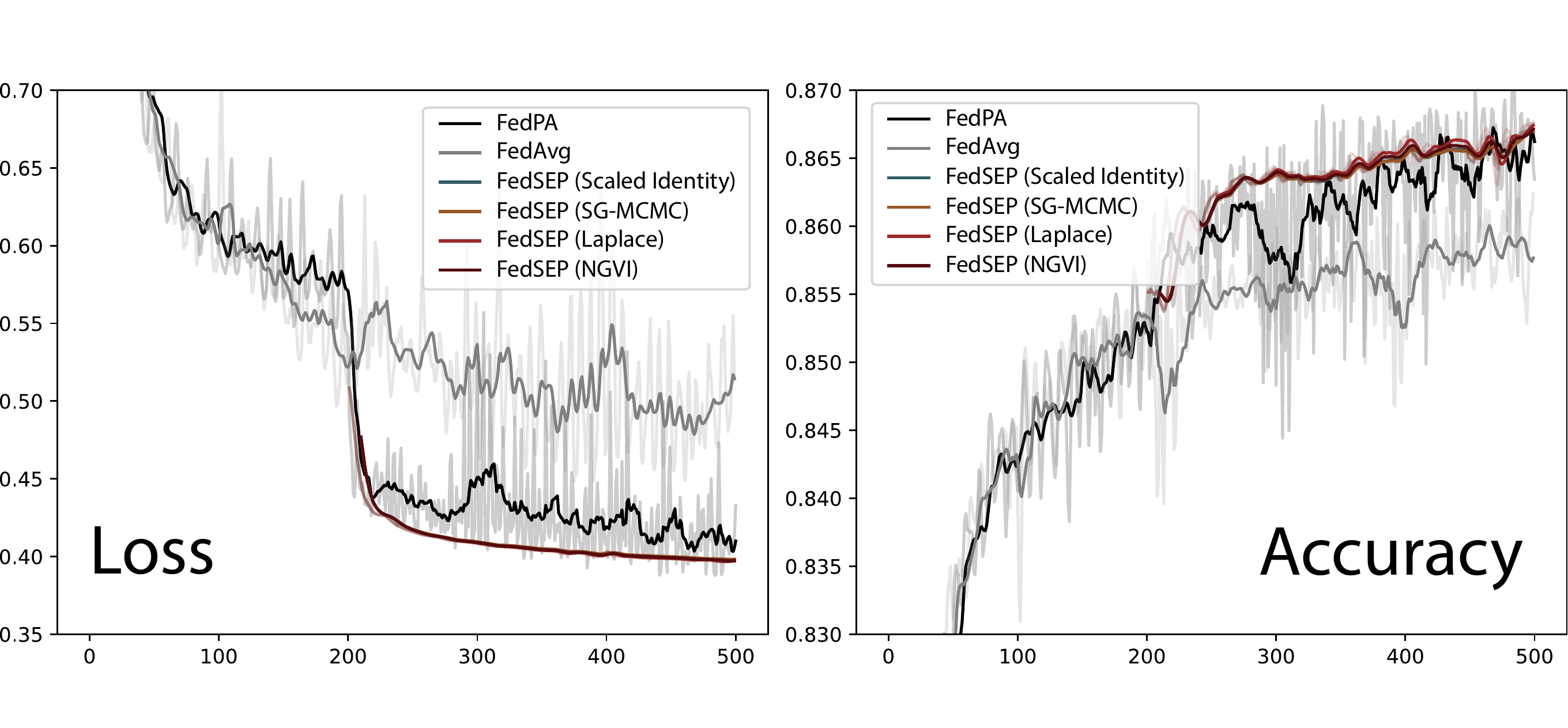}
\vspace{-17pt}
\captionof{figure}{
EMNIST-62 Experiments. Figures show the loss and accuracy of the global parameter estimation as a function of rounds for FedAvg, FedPA, and (stateless) FedSEP with various inference techniques. The transitions from FedAvg to FedPA and FedSEP happen at round $200$.}
\label{fig:emnist62-experiments}
\vspace{5pt}
\small
\centering
\begin{tabular}{l|rr|rr}
\toprule
& \multicolumn{2}{|c|}{accuracy ($\%$, $\uparrow$)} & \multicolumn{2}{c}{rounds ($\#$, $\downarrow$)} \\
\textbf { Method } & 300R & 500R & 86\% & 86.5\% \\
\midrule
FedPA                & $85.9$ & $86.5$ & $246$ & $398$ \\
FedAvg               & $85.3$ & $85.8$ & $465$ & $-$ \\
\midrule
FedSEP (I)       & $\textbf{86.1}$ & $\textbf{86.6}$ & $\textbf{228}$ & $399$ \\
FedSEP (M)           & $\textbf{86.1}$ & $\textbf{86.6}$ & $\textbf{228}$ & $399$ \\
FedSEP (L)        & $\textbf{86.1}$ & $\textbf{86.6}$ & $\textbf{228}$ & $\textbf{364}$ \\
FedSEP (V)             & $\textbf{86.1}$ & $\textbf{86.6}$ & $\textbf{228}$ & $382$ \\
\bottomrule
\end{tabular}
\captionof{table}{EMNIST-62 Experiments.
We measure the number of rounds to reach certain accuracy thresholds (based on $10$-round running averages) and the best accuracy attained within specific rounds (based on $100$-round running averages).
We use \textbf{I} (Scaled Identity Covariance), \textbf{M} (MCMC), \textbf{L} (Laplace), and \textbf{V} (NGVI) to refer to different inference techniques.
}
\label{table:emnist62-experiments}

\end{figure}

\paragraph{EMNIST.}
\label{subsec:emnist62-experiments}
Please see Fig.~\ref{fig:emnist62-experiments} and Table~\ref{table:emnist62-experiments} for experimental results.

\begin{figure}%
\centering
\includegraphics[width=0.39\linewidth]{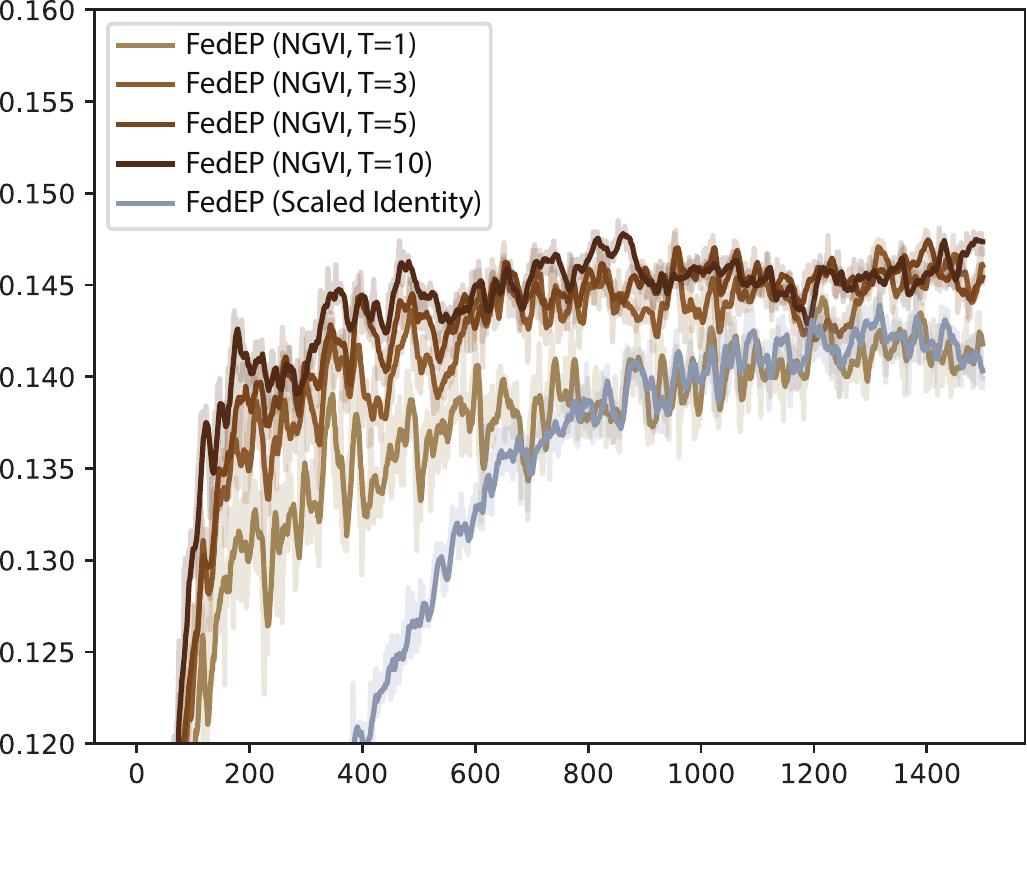}
\vspace{-15pt}
\caption{Accuracy as a function of rounds, and number of NGVI epochs ($T_{\text{NGVI}}$) in the CIFAR-100 setting, with a (relatively) small model.}
\label{fig:analysis-experiments-appendix}
\end{figure}

\paragraph{Analysis.}
\label{subsec:analysis-experiments-appendix}
This section extends the experiments (the ``small'' setting) in Sec.~\ref{subsec:analysis-experiments}. It looks at the performance as we increase the complexity (a proxy of quality) of approximate inference techniques. We vary the number of iterations in NGVI from $1$ (cheap) to $10$ (expensive) epochs. We can observe in Fig.~\ref{fig:analysis-experiments-appendix} that as we increase NGVI's computations, the performance improves.

\begin{figure}[t]
\vspace{-30pt}
\begin{algorithm}[H]
\small
\setstretch{1.70}
\caption{Approximate Inference: NGVI}
\label{alg:approximate-inference-ngvi}
\begin{algorithmic}[1]
\State \textbf{Input:} $\mathcal{D}_k, \bmu_{\backslash k}, \bSigma_{-k}, T_{\text{NGVI}}, N_{\text{NGVI}}, \beta_{\text{NGVI}}$
\State Initialize $\bolds_0, \bSigma_{\backslash k, 0}$
\For{$t = 1, \dots, T_{\text{NGVI}}$}
    \State $\mathcal{F} \leftarrow \{\}$
    \For{$i = 1, \dots, N_{\text{NGVI}}$}
        \State $\btheta \sim \mathcal{N}\left(\btheta; \bmu_{\backslash k}, \bSigma_{\backslash k, t-1}\right)$
        \State $\mathcal{F} \leftarrow \mathcal{F} \cup \frac{1}{|\mathcal{D}_k|} \operatorname{Fisher}(\btheta, \mathcal{D}_k)$.
    \EndFor
    \State $\boldF \leftarrow \operatorname{Average}(\mathcal{F})$
    \State $\bolds_t \leftarrow \beta_{\text{NGVI}} \bolds_{t-1} + \left(1 - \beta\right) \boldF$
    \State $\bSigma_{\backslash k, t} \leftarrow \left( |\mathcal{D}_k| \bolds_t + \bSigma^{-1}_{-k}  \right)^{-1}$
\EndFor
\State \textbf{Output:} $\bSigma_{\backslash k, T_{\text{NGVI}}}$
\end{algorithmic}
\end{algorithm}
\end{figure}

\subsection{Hyperparameters}
\label{subsec:hparams}
Please see Table~\ref{table:hparams} for hyperparameter details. In Table~\ref{table:cifar100-hparams}, we also conduct experiments to understand their influence on the different algorithms.

\begin{table}
\centering
\small
\begin{tabular}{l|lll}
\toprule
\textbf{Hyperparameter} & \textbf{CIFAR-100} & \textbf{StackOverflow} & \textbf{EMNIST-62}  \\
\midrule
\multicolumn{4}{c}{Task Hyperparameters from~\citet{al2020federated}}\\
\midrule
Server Optimizer           & SGD ($m=0.9$) & Adagrad ($\tau=10^{-5}$) & SGD ($m=0.9$) \\
Client Optimizer$^\dagger$ & SGD ($m=0.9$) & SGD ($m=0.9$)            & SGD ($m=0.9$) \\
Clients Per Round          & $20$          & $10$                     & $100$ \\
Server Learning Rate       & $0.5$         & $5.0$                    & $0.5$ \\
Client Learning Rate       & $0.01$        & $50.0$                   & $0.01$ \\
Client Epochs              & $10$          & $5$                      & $20$ \\
Burn In                    & $400$         & $800$                    & $200$ \\
\midrule
\multicolumn{4}{c}{Client Inference Hyperparameters}\\ 
\midrule
Scale ${\alpha_{\text{cov}}}^\ddagger$ & $5 \times 10^{-2}$ & $1 \times 10^{-8}$ & $5 \times 10^{-3}$ \\
MCMC Shrinkage                         & $1 \times 10^{-4}$ & $1 \times 10^{-6}$ & $1 \times 10^{-4}$ \\
Laplace Epochs                         & $5^\star$          & $5$                & $5$ \\
NGVI Epochs                            & $5$                & $10$               & $5$ \\
NGVI Samples                           & $5$                & $10$               & $5$ \\
NGVI $\beta_{\text{NGVI}}$             & $0.99$             & $0.99$             & $0.99$ \\
\midrule
\multicolumn{4}{c}{Client Inference Hyperparameters Search Space}\\ 
\midrule
Scale ${\alpha_{\text{cov}}}$ & $\{1, 2, 5, 10\} \times 10^{-2}$ & $1 \times \{10^{-7}, 10^{-8}, 10^{-9}\}$ & $\{1, 5\} \times \{10^{-2}, 10^{-3}, 10^{-4}\}$ \\
MCMC Shrinkage                & \multicolumn{3}{c}{$1 \times \{10^{-3}, 10^{-4}, 10^{-5}, 10^{-6}\}$}\\
Laplace Epochs                & \multicolumn{3}{c}{$\{5, 10\}$}\\
NGVI Epochs                   & \multicolumn{3}{c}{$\{5, 10\}$}\\
NGVI Samples                  & \multicolumn{3}{c}{$\{5, 10\}$}\\
NGVI $\beta_{\text{NGVI}}$    & \multicolumn{3}{c}{$\{0.9, 0.99\}$}\\
\bottomrule
\end{tabular}
\vspace{-5pt}
\caption{
Hyperparameters.
$^\dagger$Client has two separate optimizers, one used in local optimization (SG-MCMC), and one used in local state updates (for stateful FedEP). When applied, the client state optimizer reuses the same configuration as the server optimizer.
$^\ddagger$This is a per-data-point scale, and is also used in other approximate inference techniques.
$^\star$The (stateful) FedEP uses $10$ Laplace epochs. }
\label{table:hparams}
\end{table}
\begin{table}[t]
\centering
\small
\begin{tabular}{lll|ll|ll}
\toprule
&&& \multicolumn{2}{c|}{\textbf{Accuracy} ($\%$, $\uparrow$)} & \multicolumn{2}{c}{\textbf{Rounds} ($\#$, $\downarrow$)} \\
\textbf{Method}  & \textbf{Hyperparameter} & & 1000R & 1500R & 45\% & 50\% \\
\midrule
\multirow{3}{*}{FedEP (I)} & \multirow{3}{*}{Scale $\alpha_{\text{cov}}$}
&  $4 \times 10^{-2}$  & $49.1$ & $50.3$ & $457$ & $1081$ \\
&& $5 \times 10^{-2}$  & $48.9$ & $50.5$ & $464$ & $1105$ \\
&& $6 \times 10^{-2}$  & $48.8$ & $50.5$ & $474$ & $1206$ \\
\midrule
\multirow{3}{*}{FedEP (M)} & \multirow{3}{*}{MCMC Shrinkage}
&  $5 \times 10^{-5}$     & $47.8$ & $48.8$ & $482$ & $-$ \\
&& $1 \times 10^{-4}$     & $48.9$ & $50.5$ & $456$ & $1179$ \\
&& $5 \times 10^{-4}$     & $49.2$ & $49.2$ & $436$ & $-$ \\
\midrule
\multirow{2}{*}{FedEP (L)} & \multirow{2}{*}{Laplace Epochs}
&  $5 $ & $46.7$ & $47.9$ & $513$ & $-$ \\
&& $10$ & $46.7$ & $47.9$ & $514$ & $-$ \\
\midrule
\multirow{2}{*}{FedEP (V)} & \multirow{2}{*}{NGVI Epochs}
&  $5 $ & $47.9$ & $49.6$ & $478$ & $-$ \\
&& $10$ & $46.6$ & $48.5$ & $523$ & $-$ \\
\midrule
\multirow{3}{*}{FedSEP (I)} & \multirow{3}{*}{Scale $\alpha_{\text{cov}}$}
&  $4 \times 10^{-2}$ & $48.2$ & $48.7$ & $431$ & $-$ \\
&& $5 \times 10^{-2}$ & $48.3$ & $49.0$ & $431$ & $-$ \\
&& $6 \times 10^{-2}$ & $48.3$ & $49.1$ & $433$ & $-$ \\
\midrule
\multirow{3}{*}{FedSEP (M)} & \multirow{3}{*}{MCMC Shrinkage}
&  $5 \times 10^{-5}$    & $48.3$ & $48.9$ & $431$ & $-$ \\
&& $1 \times 10^{-4}$    & $48.3$ & $49.0$ & $432$ & $-$ \\
&& $5 \times 10^{-4}$    & $48.4$ & $49.0$ & $431$ & $-$ \\
\midrule
\multirow{2}{*}{FedSEP (L)} & \multirow{2}{*}{Laplace Epochs}
&  $5 $ & $47.2$ & $47.9$ & $437$ & $-$ \\
&& $10$ & $47.2$ & $47.9$ & $439$ & $-$ \\
\midrule
\multirow{2}{*}{FedSEP (V)} & \multirow{2}{*}{NGVI Epochs}
&  $5 $ & $47.9$ & $48.8$ & $432$ & $-$ \\
&& $10$ & $47.8$ & $48.6$ & $432$ & $-$ \\
\bottomrule
\end{tabular}
\captionof{table}{CIFAR-100 Hyperparameter Analysis Experiments.
FedEP and FedSEP refer to the stateful EP and stateless stochastic EP. We use \textbf{I} (Scaled Identity Covariance), \textbf{M} (MCMC), \textbf{L} (Laplace), and \textbf{V} (NGVI) to refer to different inference techniques.
}
\label{table:cifar100-hparams}
\end{table}

\end{document}